%% file: rev_3_generalizable_rl.tex
\documentclass[11pt,twoside]{article}

\usepackage{fullpage}

\input{commands}

\begin{document}

\begin{center}

{\bf{\LARGE{When Is Generalizable Reinforcement Learning Tractable?}}}



\vspace*{.2in}

{\large{
\begin{tabular}{ccc}
Dhruv Malik & Yuanzhi Li & Pradeep Ravikumar
\end{tabular}
}}
\vspace*{.2in}

\begin{tabular}{c}
Machine Learning Department \\
Carnegie Mellon University
\end{tabular}

\vspace*{.2in}

\today

\end{center}
\vspace*{.2in}

\begin{abstract}
    Agents trained by reinforcement learning (RL) often fail to generalize beyond the environment they were trained in, even when presented with new scenarios that seem similar to the training environment. We study the query complexity required to train RL agents that generalize to multiple environments. Intuitively, tractable generalization is only possible when the environments are similar or close in some sense. To capture this, we introduce \emph{\weakprox}, a natural structural condition that requires the environments to have highly similar transition and reward functions and share a policy providing optimal value. Despite such shared structure, we prove that tractable generalization is impossible in the worst case. This holds even when each individual environment can be efficiently solved to obtain an optimal linear policy, and when the agent possesses a generative model. Our lower bound applies to the more complex task of representation learning for the purpose of efficient generalization to multiple environments. On the positive side, we introduce \emph{\strprox}, a strengthened condition which we prove is sufficient for efficient generalization.
\end{abstract}



\section{Introduction}
\label{sec:intro}
Reinforcement learning (RL) is the dominant paradigm for sequential decision making in machine learning, and has achieved success in a variety of domains such as competitive gaming~\citep{mnih15, silver17} and robotic control~\citep{gu17, haarnoja18}. Despite this success, many issues prevent RL from being regularly used in the real world. For example, one typically trains and tests RL agents in the same environment. In such cases, an agent can memorize behavior that achieves high reward, without acquiring the true behavior that the system designer desires. This has raised concerns about RL agents overfitting to a single environment, instead of learning meaningful skills~\citep{farebrother18}.

Indeed, a long line of work has noted the brittleness of RL agents: slight changes in the environment, such as those incurred by modeling or simulator design errors, or slight perturbations of the agent's trajectory, can lead to catastrophic declines in performance~\citep{rajeswaran17, zhang18, henderson18}. Furthermore, although RL agents can solve difficult tasks, they struggle to transfer the skills they learned in one task to perform well in a different but similar task~\citep{rakelly19, yu19}. Yet, in the real world, it is reasonable to expect that RL agents will see scenarios that are at least mildly different from the specific scenarios they trained for.

Hence, a desirable property of RL agents is that of \emph{generalization}, broadly defined as the ability to discern the correct notion of behavior and perform well in semantically similar environments. We focus on two popular generalization settings. The \emph{Average Performance} setting assumes there is an underlying distribution over the environments that an agent might encounter. The agent's goal is to perform well on average across this distribution~\citep{packer18, nichol18, cobbe19}. The \emph{Meta Reinforcement Learning} setting is closely related~\citep{finn17, clavera18, rakelly19}. Here an agent first learns from a suite of training environments sampled from a distribution. Then at test time the agent must leverage this experience to adapt to a new environment sampled from the same distribution, via only a few queries in the new environment.

Of course, in full generality, both notions of generalization are impossible to achieve efficiently. This is especially true in the RL function approximation setting, where the \emph{cardinality} of the state space is potentially infinite, and so we desire query complexity that scales (polynomially) with the \emph{dimensionality} of the state space~\citep{du19shiftoracle, vanroy19, du20, lattimore20featurerep, weisz2021queryefficient}. Hence, key to both lines of inquiry is the premise that the environments are structurally similar. For example, a robot may face the differing tasks of screwing a bottle cap and turning a doorknob, but both tasks involve turning the wrist~\citep{rakelly19}. The hope is that if the environments are sufficiently similar, then RL can exploit this structure to efficiently discover policies that generalize.

Yet, it remains unclear what kind of structure is necessary, and what it means for different environments to be close or similar. Motivated by this, we ask the following question:
\begin{center}
\textbf{What are the structural conditions on the environments that permit efficient generalization?}
\end{center}
This question underlies the analysis of our paper. We focus on environments that share state-action spaces, since even this basic case is not well understood in the literature. Indeed, even in this simplified setting, efficient generalization can be highly non-trivial. We make the following contributions. \\

\noindent \textbf{Our Contributions.} We introduce \emph{\weakprox}, a natural structural condition that is motivated by classical RL results, and requires the environments to have highly similar transition and reward functions and share optimal trajectories. We prove a statistical lower bound demonstrating that tractable generalization is impossible, despite this shared structure. This lower bound holds even when each individual environment can be efficiently solved to obtain a linear policy providing optimal value, and when the agent possesses a generative model. Consequentially, we show that a classical metric for measuring the relative closeness of MDPs is not the right metric for modern RL generalization settings. Our lower bound implies that learning a state representation for the purpose of efficiently generalizing to multiple environments, is worst case sample inefficient --- even when such a representation exists, the environments are ostensibly similar, and any single environment can be efficiently solved.

To provide a sufficient condition for efficient generalization, we introduce \emph{\strprox{}}. This structural condition strengthens Weak Proximity by additionally constraining the environments to share an optimal policy. We provide an algorithm which exploits \strprox{} to provably and efficiently generalize, when the environments share deterministic transitions. \\

\noindent \textbf{Organization.} The remainder of this paper is organized as follows. In Section~\ref{sec:related_work}, we discuss prior work that has tackled the generalization problem in RL. In Section~\ref{sec:prob_form}, we formally present our settings of interest and state the problem that is our object of study. In Section~\ref{sec:results}, we present our theoretical results. We conclude and discuss avenues for future work in Section~\ref{sec:discussion}. All proofs are presented in Appendices~\ref{app:lower},~\ref{app:upper},~\ref{app:prop_tight} and~\ref{app:aux_proof}.

\section{Related Work}
\label{sec:related_work}
\noindent \textbf{Simulation Lemma.} Many prior works define notions of statistical distance between Markov decision processes (MDPs), and measure the relative value of policies when deployed in different MDPs that are close under such metrics. The Simulation Lemma, which uses total variation distance between transitions and the absolute difference of rewards as this metric, is a well known formalization of this and has been very useful in classical prior work~\citep{kearns99, kearns02, brafman03, kakade03, abbeel05}. These works do not directly tackle generalization, but their analyses construct an approximate MDP that models the true MDP under the aforementioned metric. Solving this approximate MDP then corresponds to solving the true MDP. It is natural to ask whether this metric is useful for measuring the similarity of MDPs in modern RL generalization settings. We show this metric is not appropriate for the settings we study. \\

\noindent \textbf{Transfer \& Multitask Learning.} There are varying formalisms of both settings, so we do not directly study them. However, they are broadly relevant, and we expect our theory to be useful for future studies of these settings. The works~\citep{lazaric10, brunskill13, jiang18, wang19} all study metrics for measuring variation between MDPs that are different from the metrics we study. A metric similar to the one used in the Simulation Lemma has also been studied~\citep{feng19}, and we show that this is inappropriate for our settings. \\

\noindent \textbf{Average Performance \& Meta RL Settings.} We directly study these two settings, which have seen much empirical work~\citep{packer18, cobbe19, cobbe19procgen, rakelly19, yu19}. On the theoretical side,~\citep{bertran20, song20} study an Average Performance setting where the agent receives a noisy observation in lieu of the actual state. We focus on the simpler setting where the agent knows its state. Recent works~\citep{fallah20, wang20meta} analyze the MAML algorithm~\citep{finn17} in the context of Meta RL. In the worst case, their complexity bounds scale exponentially with the horizon, and they do not discuss structure which permits tractable Meta RL. \\

\noindent \textbf{Representation Learning.} A large body of work has focused on extracting a representation useful for a single MDP~\citep{ferns04, castro20similar, lan21, zhang21bisim}. Some works extend this to multiple MDPs~\citep{castro10transfer, agarwal21contrastive, sonar20}, but they are about learning shared representations for MDPs that appear similar (but not from a sample efficiency perspective), while we formalize what it means for MDPs to be similar (in a sample efficient sense). Indeed, these works study the general case when the environments have distinct state spaces, but our lower bounds show generalization is non-trivial even when each MDP shares the same state space.

\section{Problem Formulation}
\label{sec:prob_form}
\textbf{Notation \& Preliminaries.} Before describing our settings of interest, we establish notation and briefly review preliminaries. We always use $M$ to denote a Markov decision process (MDP). Recall that an undiscounted finite horizon MDP is specified by a set of states $\StateSet$, a set of actions $\ActSet$, a transition function $\Trans$ which maps from state-action pairs to distributions over states, a reward function $\Rew$ which maps state-action pairs to nonnegative real numbers, and a finite planning horizon $H$. We assume that the state-action pairs are featurized, so that $\mathcal{S} \times \mathcal{A} \subset \R^d$, and that $\| (s, a) \|_2 = 1$ for all $(s, a) \in \mathcal{S} \times \mathcal{A}$. Any MDP we consider is undiscounted and has a finite action space, but could have an uncountable state space. If we need to refer to the transition or reward function of a specific MDP $M$, then we shall denote this via $\Trans_M$ or $\Rew_M$. We will denote a distribution over MDPs as $\mathcal{D}$. We also assume that $\mathcal{S}$ can be partitioned into $H$ different levels. This means that for each $s \in \mathcal{S}$ there exists a unique $h \in \{ 0, 1 \dots H-1 \}$ such that it takes $h$ timesteps to arrive at $s$ from $s_0$. We say that such a state $s$ lies on level $h$, and denote $\mathcal{S}_h$ to be the set of states on level $h$. This assumption is without loss of generality, since we can always make the final coordinate of each state-action pair encode the number of timesteps that elapsed to reach the state. A ``deterministic MDP'' is one with deterministic transitions. For any MDP, we assume a single initial state $s_0$, which strengthens our lower bounds.

A policy maps each state to a corresponding distribution over actions, and shall typically be denoted by $\pi$. The total expected reward accumulated by policy $\pi$ when initialized at state $s$ in MDP $M$ is given by $\E \left[ \sum_{h=\text{level}(s)}^{H-1} \Rew_M(s_h, a_h)~\vert~\pi \right]$ and will be denoted by $\V{s}{M}{\pi}$. Here the expectation is over the trajectory $\{ (s_h, a_h) \}_{h = \text{level}(s)}^{H-1}$ given that the first state in the trajectory is $s$. So $\V{s}{M}{\pi}$ is the value of the policy $\pi$ in MDP $M$ with respect to (w.r.t) initial state $s$. Analogously, if a policy is parameterized by $\overline{\theta} = \{ \theta_h \}_{h=0}^{H-1}$, then we denote it as $\pi(\overline{\theta})$, and the notation $\V{s}{M}{\pi}$ is then replaced by $\V{s}{M}{\overline{\theta}}$. We assume that the cumulative reward collected by any trajectory from any initial state $s$ in any MDP $M$ is always bounded by $1$. Hence the value of any policy in any MDP lies in the interval $[0,1]$. $\TvDist(P, Q)$ denotes the total variation (TV) distance between probability distributions $P$ and $Q$. Throughout, we use standard big $O$ notation.

\subsection{Problem Settings}
\noindent \textbf{Average Performance Setting.} There is a fixed distribution $\mathcal{D}$ over a family of MDPs. One can sample MDPs from $\mathcal{D}$. The algorithm can query states in the sampled MDPs, to learn some common structure. The goal is to solve
\begin{equation}
\label{eqn:avg_performance_statement}
\max_{\pi} \E_{M \sim \mathcal{D}} \left[ \V{s_0}{M}{\pi} \right].
\end{equation}

\noindent \textbf{Meta Reinforcement Learning Setting.} There is a fixed distribution $\mathcal{D}$ over a family of MDPs. At training time, one can sample MDPs from $\mathcal{D}$. The algorithm can query states in the sampled MDPs, to learn some common structure between all the MDPs. Then at test time, an MDP $\Mtest$ is sampled from the same distribution $\mathcal{D}$. The goal of the algorithm is to learn a subroutine, which with non-trivial probability over the selection of $\Mtest$, can solve
\begin{equation}
\label{eqn:meta_rl_statement}
\max_{\pi} \V{s_0}{\Mtest}{\pi},
\end{equation}
significantly more efficiently than trying to solve $\Mtest$ without having seen any training MDPs. \\

\noindent In both settings, ``sampling an MDP'' means drawing an MDP i.i.d from $\mathcal{D}$, so that the agent can then interact with it by performing trajectories in it. Note that in Eqs.~\eqref{eqn:avg_performance_statement} \&~\eqref{eqn:meta_rl_statement}, in full generality the initial state $s_0$ is random and depends on $M, \Mtest$. We focus on the case when the MDPs supporting $\mathcal{D}$ share a state-action space, and hence share the same single initial $s_0$ since we assume a single initial state for any MDP. While such assumptions are already strong, they only strengthen our lower bounds. Furthermore, it is necessary to understand this simpler setting, before looking at more complex scenarios. To the best of our knowledge, such a study has not appeared in prior work.

To solve the problems described by Eqs.~\eqref{eqn:avg_performance_statement} \&~\eqref{eqn:meta_rl_statement}, we need to define an appropriate query model for the algorithm. We consider two query models, the first of which is strictly stronger than the second. \\

\noindent \textbf{Strong Query Model (SQM).} Sampling an MDP from $\mathcal{D}$ incurs no cost. The agent has a generative model of any sampled MDP $M$. To interact with $M$, the agent inputs a state-action pair $(s,a)$ of $M$ into the model, and receives $R_M(s,a)$ and a state sampled from $\Trans_M(s,a)$. This incurs a query cost of one. The goal is to solve Eqs.~\eqref{eqn:avg_performance_statement} \&~\eqref{eqn:meta_rl_statement} with total query cost that is at most polynomial in $\vert \mathcal{A} \vert, H, d$. \\

\noindent \textbf{Weak Query Model (WQM).} Sampling an MDP from $\mathcal{D}$ incurs a query cost of $\qone \geq 1$. Within a sampled MDP $M$, the agent operates in the standard episodic RL setup. Concretely, during each episode the agent interacts with the MDP by starting from $s_0$, taking an action and observing the next state and reward, and repeating. Each action taken during an episode incurs a query cost of one. The goal is to solve Eqs.~\eqref{eqn:avg_performance_statement} \&~\eqref{eqn:meta_rl_statement} with total query cost that is at most polynomial in $\qone, \vert \mathcal{A} \vert, H, d$. \\

\noindent Note that under both SQM and WQM, we desire query cost that is polynomial in the dimension $d$ of the state-action space, as opposed to the cardinality of the state space. This is standard for our function approximation setting~\citep{du19shiftoracle, vanroy19, du20, lattimore20featurerep, weisz2021queryefficient}, since the cardinality of the state space could be infinite. Also, we separate SQM and WQM because it is well known that different query models can lead to various subtleties in analysis and sample complexity guarantees~\citep{du19shiftoracle, vanroy19, du20, lattimore20featurerep, weisz2021queryefficient}. The generative model that defines SQM assumes that we can simulate any state of our choice without performing a trajectory, which is unrealistic in practice, and is one of the strongest oracle models considered in prior literature~\citep{kearns99generative, azar12, sidford18, du20, lattimore20featurerep, agarwal20genmodel}. We shall present our lower bounds under SQM, which makes these results stronger, but shall present our upper bound under the natural and standard WQM. More generally, we emphasize that proving a \emph{lower bound} under stronger assumptions makes the lower bound stronger. By contrast, proving an \emph{upper bound} under weaker assumptions means that upper bound applies more generally.

Without any conditions on $\mathcal{D}$, the Average Performance \& Meta RL settings can be intractable, even under SQM. This will occur if the MDPs supporting $\mathcal{D}$ do not share structure. This will also occur if any individual MDP cannot be solved efficiently. Nevertheless, in practice one often deals with MDPs which share meaningful structure~\citep{cobbe19, rakelly19}. For instance, the transition distributions of the MDPs may be close in a suitable metric. Similarly, the reward functions of the MDPs might be close in an appropriate norm, or each MDP may share a set of optimal trajectories. And in practice, individual MDPs can usually be optimized efficiently~\citep{packer18, yu19}. In such cases, it is reasonable to expect tractable generalization. We are interested in formalizing conditions that permit efficient generalization. We will particularly focus on conditions which capture shared structure of the MDPs and the tractability of individual MDPs. We now formally state the problem we consider throughout our paper.
\begin{center}
\emph{Which conditions on $\mathcal{D}$ allow us to solve the Average Performance \& Meta RL settings efficiently?}
\end{center}
As mentioned above, there are two types of requirements. The first requirement should ensure that the MDPs are meaningfully similar. We formalize such conditions in Section~\ref{sec:spwp}. The second requirement should ensure that any individual MDP is efficiently solvable, else there is no hope to efficiently find policies that generalize for many MDPs. We formalize such properties in Section~\ref{sec:optcond}.

\subsection{Strong \& Weak Proximity}
\label{sec:spwp}
We now identify conditions that capture when the MDPs supporting $\mathcal{D}$ share meaningful structure. Since MDPs are defined in terms of rewards and transitions, it is very natural to impose conditions directly on the rewards and transitions. To this end, we state the following condition.

\begin{condition}[Similar Rewards \& Transitions]
\label{assump:stochastic}
The distribution $\mathcal{D}$ satisfies this condition with parameters $\rewvar, \transvar \geq 0$ when:
\begin{enumerate}[label=(\alph*), leftmargin=*]
\item Each MDP supporting $\mathcal{D}$ shares the same state-action space $\mathcal{S} \times \mathcal{A}$.
\item For all $M_i, M_j$ supporting $\mathcal{D}$ and all $(s, a) \in \StateSet \times \ActSet$ we have $\vert \Rew_{M_i}(s, a) - \Rew_{M_j}(s, a) \vert \leq \rewvar$.
\item For all $M_i, M_j$ supporting $\mathcal{D}$ and all $(s, a) \in \StateSet \times \ActSet$ we have $\TvDist(\Trans_{M_i}(s, a), \Trans_{M_j}(s, a)) \leq \transvar$.
\end{enumerate}
\end{condition}
The parameters $\rewvar, \transvar$ naturally quantify the similarity of different MDPs. Conditions of this form are canonical and have yielded fruitful research in classical RL literature~\citep{kearns99, kearns02, brafman03, kakade03, abbeel05}, in the guise of the Simulation Lemma (see Section~\ref{sec:related_work}). To concretize this condition with an example, consider a suite of simulated robotic goal reaching tasks~\citep{yu19}, where the physics simulator is the same in each task, so the transitions are fixed and $\transvar = 0$, but the goal location changes from task to task, implying that $\rewvar > 0$. We now establish our Weak Proximity condition, which strictly strengthens Condition~\ref{assump:stochastic}.

\begin{condition}[\weakprox{}]
\label{assump:lower_bound_deterministic}
The distribution $\mathcal{D}$ satisfies \weakprox{} with parameters $\rewvar, \transvar, \alpha \geq 0$ when:
\begin{enumerate}[label=(\alph*), leftmargin=*]
\item $\mathcal{D}$ satisfies Condition~\ref{assump:stochastic} with parameters $\rewvar, \transvar \geq 0$.
\item There exists a deterministic policy $\pi^\star$ which for any MDP $M$ satisfies
$$
\V{s_0}{M}{\pi^\star} \geq \max_{\pi'} \V{s_0}{M}{\pi'} - \alpha.
$$
\end{enumerate}
\end{condition}
Weak Proximity strengthens Condition~\ref{assump:stochastic} by additionally requiring (via part (b)) that there exists some policy $\pi^\star$ which provides $\alpha$-suboptimal value for each MDP supporting $\mathcal{D}$. Intuitively, this condition implicitly constrains the MDPs to be similar, since there is a single policy which provides (nearly) optimal value, irrespective of the MDP it is deployed in. Furthermore, recall from Eqs.~\eqref{eqn:avg_performance_statement} \&~\eqref{eqn:meta_rl_statement} that the objectives of the Average Performance \& Meta RL settings are defined in terms of value w.r.t the initial state $s_0$. So it is natural to assume, as we do in part (b), that there is one policy which provides good value w.r.t $s_0$ for all MDPs. From an algorithmic perspective, this is helpful, because it ensures that we can restrict our search to those policies which perform well for many MDPs supporting $\mathcal{D}$.

Although Condition~\ref{assump:stochastic} is natural and well motivated by classical RL literature, it (and Weak Proximity) may seem strong. This is because it requires that each MDP supporting $\mathcal{D}$ shares the same state space, which may not hold in practice. We stress that we will prove a lower bound under Weak Proximity, showing that efficient generalization is impossible even in the simpler regime of a shared state space.

We now present Strong Proximity, a condition which strictly strengthens Weak Proximity. We will later show that unlike its Weak counterpart, Strong Proximity indeed permits efficient generalization. 

\begin{condition}[\strprox{}]
\label{assump:upper_bound}
The distribution $\mathcal{D}$ satisfies \strprox{} with parameters $\rewvar, \transvar, \alpha \geq 0$ when:
\begin{enumerate}[label=(\alph*), leftmargin=*]
\item $\mathcal{D}$ satisfies Condition~\ref{assump:stochastic} with parameters $\rewvar, \transvar \geq 0$.
\item There exists a deterministic policy $\pi^\star$ which is a near optimal policy for each MDP. Concretely, the policy $\pi^\star$ satisfies
$$
\V{s}{M}{\pi^\star} \geq \max_{\pi'} \V{s}{M}{\pi'} - \polsub,
$$
for each state $s$ and each MDP $M$.
\end{enumerate}
\end{condition}

\noindent Let us compare Weak with Strong Proximity. Part (a) remains identical. But \weakprox{} (b) only requires a shared policy which provides $\alpha$-suboptimal value with respect to $s_0$. This is in contrast to the shared policy in part (b) of \strprox{}, which provides $\alpha$-suboptimal value for \emph{any} state.

\subsection{Tractability of Individual Optimization}
\label{sec:optcond}
As discussed previously, in order to efficiently solve Eqs.~\eqref{eqn:avg_performance_statement} \&~\eqref{eqn:meta_rl_statement}, we require the property that each individual MDP supporting $\mathcal{D}$ can be efficiently solved. It is natural to expect such a property to hold in practice. For instance, in the context of our earlier example of simulated robotic goal reaching tasks~\citep{yu19}, any individual task can be efficiently solved via policy gradient methods. We now state two such properties, the first of which is strictly stronger than the second. Since these properties require a notion of query cost, we state both of them with reference to a generic query model QM, and when we later present our results we will instantiate QM to be either SQM or WQM. To avoid complicating notation in these statements, we assume in this subsection (as is our focus throughout the paper) that all MDPs supporting $\mathcal{D}$ are defined on the same state-action space $\mathcal{S} \times \mathcal{A} \subset \R^d$. Recall that a linear policy $\pi$ is parameterized by $\overline{\theta} = \{ \theta_h \}_{h=0}^{H-1}$, where $\theta_h \in \R^d$ and $\| \theta_h \|_2 = 1$ for all $0 \leq h \leq H-1$, such that $\pi(s) \in \argmax_{a \in \mathcal{A}} (s, a)^T \theta_h$ for any $s \in \mathcal{S}_h$. Here $x^Ty$ denotes the Euclidean inner product of $x,y \in \R^d$. We use $\pi_M^\star$ to denote an arbitrary deterministic optimal policy of MDP $M$.


\begin{property}[Strong Individual Optimization (SIO)]
Let the query model be QM. The distribution $\mathcal{D}$ satisfies SIO with parameters $k > 0$ and $0 \leq \beta < \nicefrac{1}{4}$ when:
\begin{enumerate}[label=(\alph*), leftmargin=*]
\item Any MDP $M$ supporting $\mathcal{D}$ admits an optimal linear policy. Concretely, given any $M$, there exists $\overline{\theta^\star} = \{ \theta_h^\star \}_{h=0}^{H-1}$ such that for every state $s \in \mathcal{S}_h$ we have
$$
\pi^\star_M(s) \in \argmax_{a \in \mathcal{A} } (s, a)^T \theta_h^\star.
$$
\item There exists a fixed and known algorithm, such that given any MDP $M$ and any state $s$, this algorithm uses at most $\mathcal{O}(\vert \ActSet \vert H^{k})$ query cost (under QM) on $M$ to identify (almost surely) a linear policy $\pi(\overline{\theta})$ parameterized by $\overline{\theta} = \{ \theta_h \}_{h=0}^{H-1}$ which satisfies
$$
\max_{\pi'} \V{s}{M}{\pi'} \geq \V{s}{M}{\overline{\theta}} \geq \max_{\pi'} \V{s}{M}{\pi'} - \beta.
$$
This algorithm then outputs $\pi(\overline{\theta})$ as well as $\V{s}{M}{\overline{\theta}}$.
\end{enumerate}
\end{property}
Let us discuss this property. Part (a) requires that for any MDP supporting $\mathcal{D}$, there exists an optimal linear policy. Part (b) requires that the user has knowledge of an algorithm, which can efficiently find a linear policy providing $\beta$-suboptimal value from any input state $s$ in any MDP $M$. The exponent $k$ describes the (polynomially sized) complexity of this algorithm.

SIO is a fairly strong property, since it says that a linear policy is sufficient to optimize any individual MDP, whereas in practice one typically requires nonlinear neural network policies. SIO also heavily constrains each individual MDP supporting $\mathcal{D}$ to be efficiently solvable from any initial state. We stress that we will prove our \emph{lower bounds} under SIO, which makes our result stronger. Meanwhile, we prove our \emph{upper bounds} under the following property, which is significantly weaker than SIO.

\begin{property}[Weak Individual Optimization (WIO)]
Let the query model be QM. The distribution $\mathcal{D}$ satisfies WIO with parameter $0 \leq \beta < \nicefrac{1}{4}$ when the following holds. There exists an oracle $\oracle$, which takes as input a state $s$ and MDP $M$, and outputs $\approxV{s}{M}$ satisfying
$$
\max_{\pi'} \V{s}{M}{\pi'} \geq \approxV{s}{M} \geq \max_{\pi'} \V{s}{M}{\pi'} - \solvesub,
$$
via query cost (under QM) on $M$ that is polynomial in $\vert \ActSet \vert, H, d$.
\end{property}
WIO postulates the existence of an oracle $\oracle$, which can efficiently approximate the optimal value that is achievable from an input state and MDP. To see that WIO is strictly weaker than SIO, simply note we can implement $\oracle$ by running the algorithm described in part (b) of SIO. Note that in certain states, a user may use domain knowledge to implement $\oracle$ without solving an entire RL problem. Also note that WIO does not place (arguably unrealistic) linearity restrictions on the MDPs supporting $\mathcal{D}$. Finally, note that although the algorithm in SIO (b) returns both a policy and a scalar value, the oracle in WIO only returns a scalar value.

\section{Main Results}
\label{sec:results}
We shall present our results in two subsections. In Section~\ref{sec:lower_bounds}, we prove lower bounds which demonstrate that even under Weak Proximity, SQM and SIO, tractable generalization is worst case impossible. In Section~\ref{sec:upper_bound}, we prove that efficient generalization is possible under Strong Proximity, WQM and WIO, when the MDPs supporting $\mathcal{D}$ share a deterministic transition function.

\subsection{Lower Bounds}
\label{sec:lower_bounds}
Before stating our own results, we first state the following classical result which is known as the Simulation Lemma~\citep{kearns99, kearns02, brafman03, kakade03, abbeel05}. Recall that $\rewvar, \transvar$ are parameters used to satisfy Condition~\ref{assump:stochastic}.

\begin{lemma}
\label{lem:simulation}
Consider any $\mathcal{D}$ satisfying Condition~\ref{assump:stochastic} with $\rewvar, \transvar \geq 0$. For any policy $\pi$ and any $M_1, M_2$ supporting $\mathcal{D}$, we have that $\vert \V{s_0}{M_1}{\pi} - \V{s_0}{M_2}{\pi} \vert \leq \rewvar H + \transvar H$.
\end{lemma}
This result is almost identical to the one given by~\citep{kakade03}, although there are some (minor) differences in assumptions so we provide a proof in Appendix~\ref{app:aux_proof}. This lemma shows that when $\mathcal{D}$ satisfies Condition~\ref{assump:stochastic} and $\rewvar, \transvar$ are each $o(\frac{1}{H})$, then efficient generalization is trivial, at least in problems where $H$ is large and we want to optimize to within $o(1)$ tolerance. Concretely, take any $M$ supporting $\mathcal{D}$ and use a standard RL method to find $\pi$ which satisfies $\V{s_0}{M}{\pi} \approx \max_{\pi'} \V{s_0}{M}{\pi'}$. Then Lemma~\ref{lem:simulation} ensures $\V{s_0}{M'}{\pi} \gtrsim \max_{\pi'} \V{s_0}{M'}{\pi'} - o(1)$ for any other MDP $M'$ supporting $\mathcal{D}$. This implies $\E_{M \sim \mathcal{D}} \left[ \V{s_0}{M}{\pi} \right] \gtrsim \max_{\pi'} \E_{M \sim \mathcal{D}} \left[ \V{s_0}{M}{\pi'} \right] - o(1)$ and $\V{s_0}{\Mtest}{\pi} \gtrsim \max_{\pi'} \V{s_0}{\Mtest}{\pi'} - o(1)$.

Since \weakprox{} implies Condition~\ref{assump:stochastic}, Lemma~\ref{lem:simulation} and all the above statements remain true when $\mathcal{D}$ satisfies \weakprox{}. Naturally then, in our settings it is only interesting to consider problems when at least one of either $\rewvar$ or $\transvar$ is $\Omega(\frac{1}{H})$. Our next result is a \emph{lower bound} which shows that when $\rewvar = \Theta(\frac{1}{H})$ and $\transvar = 0$, then \weakprox{} is not sufficient to efficiently generalize in the Average Performance Setting. For the statement of this result, recall that $\rewvar, \transvar, \alpha$ are parameters used to satisfy \weakprox{}, while $\beta, k$ are parameters used to satisfy SIO.

\begin{theorem}
\label{thm:avg_per_det_lower_bound}
Let the query model be SQM. For any $k \geq 3$, there exists $\mathcal{D}$ satisfying \weakprox{} with $\rewvar =\Theta(\frac{1}{H})$, $\transvar = 0$ \& $\alpha = 0$ and SIO with $\beta = 0$ \& $k$, such that the MDPs supporting $\mathcal{D}$ are deterministic and the following holds. Any (possibly randomized) algorithm requires $\Omega \left( \min \left \{ \vert \ActSet \vert^H, 2^d \right \} \right)$ total query cost to find (with probability at least $\nicefrac{1}{2}$ over the randomness of the algorithm) a policy $\pi$ satisfying
\begin{equation*}
\E_{M \sim \mathcal{D}} \left[ \V{s_0}{M}{\pi} \right] \geq \max_{\emph{\text{linear policy }} \pi'} \E_{M \sim \mathcal{D}} \left[ \V{s_0}{M}{\pi'} \right] - \nicefrac{1}{4}.
\end{equation*}
\end{theorem}

\noindent We defer the proof to Appendix~\ref{app:lower_thm}. Let us discuss this theorem, which is stated for the Average Performance Setting, when the MDPs supporting $\mathcal{D}$ all share a deterministic transition function. Recall that SQM is the stronger query model we consider, which strengthens this lower bound, and trivially implies a lower bound for when WQM is the query model. Also recall that SIO is the stronger individual optimization property that we consider, and it ensures that the user can efficiently find a linear policy providing optimal value w.r.t any initial state for any individual MDP, since $\beta = 0$. Moreover, \weakprox{} (b) ensures that each MDP supporting $\mathcal{D}$ shares a policy that provides optimal value (w.r.t $s_0$), since $\alpha = 0$. And \weakprox{} (a) \emph{explicitly} requires that the reward functions are (non-trivially) close, in the sense defined by Condition~\ref{assump:stochastic}, because $\rewvar = \Theta(\frac{1}{H})$. Despite this significant structure, the theorem demonstrates that one can still require an exponential query cost to find a policy that is nearly as good as the best \emph{linear} policy (which is of course easier than finding the best generic policy). Note that this lower bound holds with $\alpha=\beta=\transvar=0$, and so implies a lower bound for when any of $\alpha,\beta,\transvar$ are strictly positive. As we discuss at the end of Section~\ref{sec:lower_bounds}, Theorem~\ref{thm:avg_per_det_lower_bound} (and its forthcoming corollaries) immediately applies to the task of learning a feature mapping which maps similar states to the same vector, for the purpose of efficiently solving Average Performance and Meta RL settings.

We note that in the construction used to prove the lower bound of Theorem~\ref{thm:avg_per_det_lower_bound}, the algorithm we provide to satisfy the SIO property is extremely simple and natural. It is simply a greedy version of Monte Carlo Tree Search, which is extremely popular in practice~\citep{kocsis06, silver17}.

Let us provide some intuition for our proof of Theorem~\ref{thm:avg_per_det_lower_bound}. In the $\vert \mathcal{A} \vert$-ary tree hard instance used in our proof, there are $\Omega(\vert \ActSet \vert^H)$ possible trajectories. The fact that $\rewvar=\Theta(\frac{1}{H})$ allows us enough degrees of freedom to hide the policy that generalizes across $\mathcal{D}$, so that identifying it requires querying each of the $\Omega(\vert \ActSet \vert^H)$ trajectories. We leverage recent techniques~\citep{du20, wang20} to construct a suitable featurization of the state-action space, that is expressive enough to allow for efficiently finding an optimal linear policy for any single MDP, but does not leak any further information. Notably, our constructed featurization satisfies that $d$ is a polynomial function of $H$.

A similar result holds for the Meta RL setting. Recall that by SIO (b), the user has access to an algorithm which can solve any $\Mtest$ at test time in $\mathcal{O}(\vert \ActSet \vert H^{k})$ queries, \emph{even if it does no training}. So it only makes sense to train, if one can use this training to solve $\Mtest$ in $o(\vert \ActSet \vert H^k)$ queries. The following corollary to Theorem~\ref{thm:avg_per_det_lower_bound} demonstrates that this may require exponential query cost during training time. Its proof is presented in Appendix~\ref{app:lower_corr_meta_det}.
\begin{corollary}
\label{corollary:meta_rl_det}
Let the query model be SQM. For any $k \geq 3$, there exists $\mathcal{D}$ satisfying \weakprox{} with $\rewvar =\Theta(\frac{1}{H})$, $\transvar = 0$ \& $\alpha = 0$ and SIO with $\beta = 0$ \& $k$, such that the MDPs supporting $\mathcal{D}$ are deterministic and the following holds. If a (possibly randomized) algorithm at test time can identify $\pi$ satisfying
\begin{equation*}
\V{s_0}{\Mtest}{\pi} \geq \max_{\emph{\text{linear policy }} \pi'} \V{s_0}{\Mtest}{\pi'} - \nicefrac{1}{4}
\end{equation*}
in $o(\vert \ActSet \vert H^k)$ queries, with probability at least $\nicefrac{1}{2}$ over the selection of $\Mtest$ (and the randomness of the algorithm), then this algorithm must have required $\Omega \left( \min \left \{ \vert \ActSet \vert^H, 2^d \right \} \right)$ total query cost at training time.
\end{corollary}

\noindent So far we presented results for when the MDPs supporting $\mathcal{D}$ share deterministic transitions but have (slightly) varying rewards. For the remainder of Section~\ref{sec:lower_bounds}, we present analogous results for when the MDPs share a reward function but have (slightly) varying transitions, again under both SIO and SQM. Recall from our discussion of Lemma~\ref{lem:simulation} that when $\rewvar = 0$, it is only interesting to consider problems when $\transvar$ is $\Omega(\frac{1}{H})$. Unfortunately, the following corollary of Theorem~\ref{thm:avg_per_det_lower_bound} shows that efficiently solving the Average Performance Setting is impossible in this regime.

\begin{corollary}
\label{corollary:avg_per_stoch}
Let the query model be SQM. For any $k \geq 3$, there exists $\mathcal{D}$ satisfying \weakprox{} with $\rewvar =0$, $\transvar = \Theta(\frac{1}{H})$ \& $\alpha = 0$ and SIO with $\beta = 0$ \& $k$, such that the following holds. Any (possibly randomized) algorithm requires $\Omega \left( \min \left \{ \vert \ActSet \vert^H, 2^d \right \} \right)$ total query cost to find (with probability at least $\nicefrac{1}{2}$ over the randomness of the algorithm) a policy $\pi$ satisfying
\begin{equation*}
\E_{M \sim \mathcal{D}} \left[ \V{s_0}{M}{\pi} \right] \geq \max_{\emph{\text{linear policy }} \pi'} \E_{M \sim \mathcal{D}} \left[ \V{s_0}{M}{\pi'} \right] - \nicefrac{1}{4}.
\end{equation*}
\end{corollary}
We defer the proof to Appendix~\ref{app:lower_corr_avg_per_stoch}. Recall the discussion of Theorem~\ref{thm:avg_per_det_lower_bound}, and note that the same discussion applies here, after swapping $\transvar$ with $\rewvar$. An analogous result holds for the Meta RL setting. As we discussed before presenting Corollary~\ref{corollary:meta_rl_det}, it only makes sense to train, if one can use this training in order to solve $\Mtest$ in $o(\vert \ActSet \vert H^k)$ queries. The following result shows that this is impossible without exponential query cost at training time. Its proof is presented in Appendix~\ref{app:lower_corr_meta_stoch}.\begin{corollary}
\label{corollary:meta_rl_stoch}
Let the query model be SQM. For any $k \geq 3$, there exists $\mathcal{D}$ satisfying \weakprox{} with $\rewvar =0$, $\transvar = \Theta(\frac{1}{H})$ \& $\alpha = 0$ and SIO with $\beta = 0$ \& $k$, such that the following holds. If a (possibly randomized) algorithm at test time can identify $\pi$ satisfying
\begin{equation*}
\V{s_0}{\Mtest}{\pi} \geq \max_{\emph{\text{linear policy }} \pi'} \V{s_0}{\Mtest}{\pi'} - \nicefrac{1}{4}
\end{equation*}
in $o(\vert \ActSet \vert H^k)$ queries, with probability at least $\nicefrac{1}{2}$ over the selection of $\Mtest$ (and the randomness of the algorithm), then this algorithm must have required $\Omega \left( \min \left \{ \vert \ActSet \vert^H, 2^d \right \} \right)$ total query cost at training time.
\end{corollary}

\noindent In conjunction with Lemma~\ref{lem:simulation}, the results of Theorem~\ref{thm:avg_per_det_lower_bound} and Corollaries~\ref{corollary:meta_rl_det},~\ref{corollary:avg_per_stoch} \&~\ref{corollary:meta_rl_stoch} suggest that the classical (and natural) way of measuring variation in MDPs using Condition~\ref{assump:stochastic} is inappropriate for the modern Average Performance \& Meta RL settings. When both $\rewvar$ and $\transvar$ are $o(\frac{1}{H})$, then these settings are trivially solvable. But when either $\rewvar$ or $\transvar$ is $\Theta(\frac{1}{H})$ then these settings become exponentially hard, even with the additional \weakprox{} condition as well as SIO \& SQM.

Theorem~\ref{thm:avg_per_det_lower_bound} and Corollaries~\ref{corollary:meta_rl_det},~\ref{corollary:avg_per_stoch} \&~\ref{corollary:meta_rl_stoch} all hold in the setting where each MDP supporting $\mathcal{D}$ shares a state-action space. So these lower bounds immediately apply to more complex settings where the MDPs are defined on disjoint state-action spaces, and where learning an appropriate representation is necessary. Indeed, it is popular in practice to learn a feature mapping which maps similar states to the same vector. Our results show that if such a mapping enables efficient solution of the Average Performance \& Meta RL settings, then learning the mapping itself is worst case inefficient.

\subsection{Upper Bound}
\label{sec:upper_bound}
We now show that Strong Proximity permits efficient generalization when the MDPs supporting $\mathcal{D}$ share deterministic transitions. While this setting is restricted, we study it because our Theorem~\ref{thm:avg_per_det_lower_bound} shows that even this setting can be worst case inefficient under strong assumptions. Furthermore, past literature on even traditional RL with a single MDP has often focused on the deterministic setting~\citep{zheng13detfunc, du20detfunc}. Notably, to prove our upper bound we only require the weaker WQM and weaker WIO. Our method is defined in Algorithm~\ref{alg:main}. It exploits Strong Proximity, which requires the existence of a policy which provides optimal value for each MDP from \emph{any} given initial state, even though the objectives in Eqs.~\eqref{eqn:avg_performance_statement} \& \eqref{eqn:meta_rl_statement} are defined only in terms of value w.r.t $s_0$.

\begin{algorithm}
\caption{GenRL}
\label{alg:main}
\begin{algorithmic}[1]
\STATE \text{Inputs: horizon length $H$, distribution $\mathcal{D}$, sample size $n$, oracle $\oracle$ as defined in WIO}
\STATE Initialize $\pi$ as an arbitrary function from $\{ 0, 1 \dots H - 1 \}$ to $\mathcal{A}$
\FOR{$t \in \{0, 1 \dots H-1 \}$}
\FOR{$i \in \{ 1, 2 \dots n \}$}
\STATE Sample $M_i \sim \mathcal{D}$
\FOR{$a \in \ActSet$}
\STATE Begin a new episode in $M_i$ at $s_0$
\IF{$t>0$}
\STATE Execute action sequence $\{ \pi(t') \}_{0 \leq t' < t}$ to arrive at $s_t$
\ENDIF
\STATE Take action $a$ to arrive at $s' = \mathcal{T}_{M_i}(s_t, a)$ and receive $\Rew_{M_i}(s_t, a)$
\STATE Query $\oracle$ to obtain $\approxV{s'}{M_i}$ and store $Q_{i, a} = \Rew_{M_i}(s_t, a) + \approxV{s'}{M_i}$
\ENDFOR
\ENDFOR

\STATE Store $a_t \in \argmax_{a' \in \ActSet} \{ \frac{1}{n} \sum_{i=1}^n Q_{i, a'} \}$ and define $\pi(t) = a_t$
\ENDFOR
\STATE \textbf{return } $\pi$
\end{algorithmic}
\end{algorithm}

\noindent Let us describe Algorithm~\ref{alg:main}. It represents policy $\pi$ as a vector which stores one action for each timestep in $\{0, 1 \dots H-1 \}$. It initializes arbitrary $\pi$ and incrementally updates it at each timestep $t$. At the beginning of any timestep $t > 0$, $\pi$ has been constructed to play the action $\pi(t') = a_{t'}$ at each timestep $t' < t$. The algorithm then executes $\{ \pi(t') \}_{0 \leq t' < t}$ to arrive at $s_t$. Crucially, due to the assumption of a shared state-action space and shared deterministic transitions, the state $s_t$ is fully determined by $\pi$ and does \emph{not} depend on the particular $M_i$. Exploiting WIO, the method queries $\oracle$ to estimate the value in $M_i$ of each child state of $s_t$. Averaging this estimated value over $\{ M_i \}_{i=1}^n$ yields an estimate of the expected value (over the randomness in $\mathcal{D}$) of each action at $s_t$. Finally, the algorithm picks the action $a_t$ with the highest estimated value, and updates $\pi$ to play $a_t$ at timestep $t$. This algorithm operates in the standard RL framework and falls under the purview of WQM.

The following result provides a performance guarantee for Algorithm~\ref{alg:main}. Recall that $\alpha, \transvar, \rewvar$ are parameters used to satisfy Strong Proximity and $\beta$ is a parameter used to satisfy WIO.

\begin{theorem}
\label{thm:upper_bound}
Let the query model be WQM. Consider any $\mathcal{D}$ satisfying WIO with $\beta \geq 0$ and \strprox{} with $\transvar = 0$ and any $\polsub, \rewvar \geq 0$, such that the MDPs supporting $\mathcal{D}$ are deterministic. Fix $\epsilon, \delta > 0$, and let $\pi$ be the output of Algorithm~\ref{alg:main} when run with $n = \frac{H^2}{\epsilon^2} \log \left( \frac{2 H \vert \ActSet \vert}{\delta} \right)$ samples. Then with probability at least $1 - \delta$, we are guaranteed that
\begin{equation*}
\E_{M \sim \mathcal{D}} \left[ \V{s_0}{M}{\pi} \right] \geq \max_{\pi'} \E_{M \sim \mathcal{D}} \left[ \V{s_0}{M}{\pi'} \right] - \epsilon - 3 \polsub H - 3 \solvesub H.
\end{equation*}
Hence the total query cost under WQM required to achieve this guarantee is polynomial in $\qone, \vert \mathcal{A} \vert, H, d$.
\end{theorem}

\noindent We defer the proof to Appendix~\ref{app:upper}. A few comments are in order. First, note that Theorem~\ref{thm:upper_bound} directly provides a guarantee for the Average Performance setting. It also provides a guarantee for the Meta RL setting, since the $\pi$ found by Algorithm~\ref{alg:main} will on average perform well for $\Mtest$, and the user can use $\pi$ to warm start any finetuning or adaptation at test time. Second, the specified value of $n$ depends only on quantities that are either known a priori or chosen by the user. This makes Algorithm~\ref{alg:main} \emph{parameter free} --- the user does not need to know the values of $\polsub, \solvesub, \rewvar, \transvar$ to run this method.

Third, note Theorem~\ref{thm:upper_bound} holds under WIO. By contrast, Weak Proximity was insufficient for efficient generalization even when paired with SIO. This suggests that a condition that is both necessary and sufficient for efficient generalization lies somewhere between Weak and Strong Proximity --- assuming, of course, that we do not assume an individual optimization property that is even stronger than SIO. Indeed, SIO is already quite strong, since SIO says that a linear policy is sufficient to optimize any individual MDP, but in practice one typically employs nonlinear neural network policies.

Finally, observe that $\rewvar$ does not appear in the error bound. So $\rewvar$ can be arbitrarily large, and Theorem~\ref{thm:upper_bound} requires no \emph{explicit} conditions on the reward functions of the MDPs supporting $\mathcal{D}$, as in the sense of Condition~\ref{assump:stochastic}. Instead, the \emph{implicit} reward structure induced by the shared nearly optimal policy required by \strprox{} is sufficient. Comparing this observation with the result of Theorem~\ref{thm:avg_per_det_lower_bound} suggests that the classical explicit constraints on rewards and transitions is not appropriate for modern RL generalization settings. Instead, implicit constraints of the sort afforded by Strong Proximity offer a more fine grained characterization of when efficient generalization is possible.

Recall from Theorem~\ref{thm:avg_per_det_lower_bound} that a lower bound holds for Weak Proximity and SIO even with $\alpha = \beta = 0$. However, Strong Proximity and WIO provide enough structure that the error bound of Theorem~\ref{thm:upper_bound} can tolerate $\alpha, \beta \geq 0$. But these $\polsub, \solvesub$ terms in the error bound of Theorem~\ref{thm:upper_bound} scale linearly with $H$. It is natural to question whether this scaling is due to a suboptimality of Algorithm~\ref{alg:main} or looseness in our analysis. We provide a partial answer to this question in Appendix~\ref{app:prop_tight}, where we prove that the dependency on $\solvesub$ given in the result of Theorem~\ref{thm:upper_bound} is tight to within a logarithmic factor in $H$.

\section{Discussion}
\label{sec:discussion}
In this paper, we studied the design of RL agents that generalize. We proved that efficient generalization is worst case impossible, even under structural conditions like \weakprox{} and strong assumptions on the query model and tractability of individual MDPs. This result extends to the task of learning representations for the purpose of efficient generalization. On the positive side, we provided Strong Proximity, which permits efficient generalization, even under mild assumptions on the query model and individual tractability. Our analysis highlights that classical metrics for measuring similarity of MDPs are inappropriate for modern RL. It also suggests that a condition which is both necessary and sufficient for efficient generalization lies between Weak \& Strong Proximity --- unless we make (arguably unreasonable) assumptions on the tractability of individual MDPs.

The primary limitation of our work is that our upper bound has limited applicability. It holds only when the MDPs share a state-action space, and when the MDPs are deterministic, which is very restrictive in practice. Our rationale for working in this restricted setting was due to our lower bounds, which show that even this toy setting can be worst case inefficient, and because it is necessary to understand the toy setting before looking at more complex scenarios. Nevertheless, our upper bound is several steps removed from the practice of RL. It is best interpreted as a preliminary sufficient condition for when efficient generalization is possible, albeit in a toy setting, and is far from conclusive on this matter.

Note that our upper bound might apply if we are a priori given a feature mapping which maps similar states of different MDPs to the same state space. For example, in self driving, learning to drive in different countries might be difficult because the images of traffic signs are different. But if a known feature map extracts the underlying meaning of these signs, then our upper bound could conceivably apply. Of course, such a known feature map is rarely available a priori, and is usually learned from data. The key direction for future work, is how to learn such a feature mapping efficiently, while ensuring that it is still useful for generalization.

\subsection*{Acknowledgements}
The authors thank Ruosong Wang and Adarsh Prasad for enlightening discussions and insightful commentary on the nature of the results. The authors also thank Ojash Neopane, Bingbin Liu, Vishwak Srinivasan, Ashwini Pokle, Stephanie Milani and Tanya Marwah for editing an earlier draft of this paper. This material is based upon work supported by the National Science Foundation Graduate Research Fellowship Program under Grant No. DGE1745016. Any opinions, findings, and conclusions or recommendations expressed in this material are those of the authors and do not necessarily reflect the views of the National Science Foundation.

\appendix
\section{Lower Bound Proof Details}
\label{app:lower}
In this section, we will provide proofs of Theorem~\ref{thm:avg_per_det_lower_bound} and Corollaries~\ref{corollary:meta_rl_det},~\ref{corollary:avg_per_stoch} \&~\ref{corollary:meta_rl_stoch}. For ease in presentation, we shall assume throughout that the action space $\ActSet$ for each MDP contains two actions, which we denote $a_1$ and $a_2$. It is easy to extend the proofs to the case when there are many actions. We will often use the notation $s_1$ and $s_2$ to denote the child states of a state $s$ when taking actions $a_1$ and $a_2$ respectively. We shall also use $\pi^*_M$ to denote an optimal policy for MDP $M$. Whenever we require $H$ to be sufficiently large for an algebraic argument to go through, we shall assume so.

\subsection{Proof Of Theorem~\ref{thm:avg_per_det_lower_bound}}
\label{app:lower_thm}

\begin{figure}[!tbp]
\centering
\scalebox{1}{
\begin{tikzpicture}[->,shorten <=5pt,shorten >=5pt,level distance=1cm]
\tikzset{}
\Tree [.\node[draw,circle] {\(s_{0}\)}; [.\node[draw,circle] {}; [.\node[draw,circle] {}; [.\(\ldots\) [.\node[draw,circle] {}; ] [.\node[draw,circle] {}; ] ] [.\(\ldots\) [.\node[draw,circle] {}; ] [.\node[draw,circle] {}; ] ] ] [.\node[draw,circle] {}; [.\(\ldots\) [.\node[draw,circle] {}; ] [.\node[draw,circle] {}; ] ] [.\(\ldots\) [.\node[draw,circle] {}; ] [.\node[draw,circle] {}; ] ] ] ]
\edge[transparent] node{}; [.{} \edge[transparent] node{}; [.{} \edge[transparent] node{}; [.{} \edge[transparent] node{}; [.{\(\ldots\)} ] ] ] ]
\edge[] node{}; [.\node[draw,circle] {}; [.\node[draw,circle] {}; [.\(\ldots\) [.\node[draw,circle] {}; ] [.\node[draw,circle] {}; ] ] [.\(\ldots\) [.\node[draw,circle] {}; ] [.\node[draw,circle] {}; ] ] ] [.\node[draw,circle] {}; [.\(\ldots\) [.\node[draw,circle] {}; ] [.\node[draw,circle] {}; ] ] [.\(\ldots\) [.\node[draw,circle] {}; ] [.\node[draw,circle] {}; ] ] ] ] ]
\end{tikzpicture}
}
\caption{An illustration of the generic binary tree structure used to define the MDPs that support the $\mathcal{D}$ constructed in the proof of Theorem~\ref{thm:avg_per_det_lower_bound}.}
\label{fig:lower_det}
\end{figure}
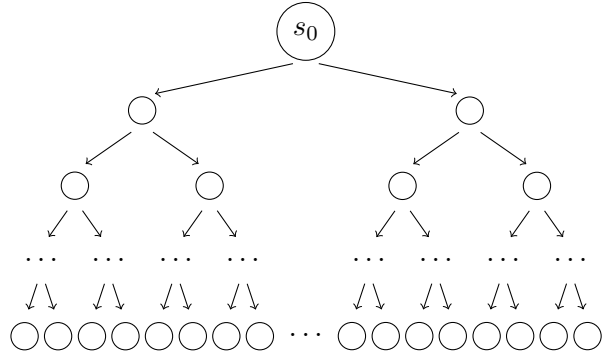

\textbf{Construction Details.} Our construction of $\mathcal{D}$ will consist of MDPs whose shared state and action spaces are generically defined by a binary tree. The structure of the binary tree is depicted in Figure~\ref{fig:lower_det}. The tree is of length $H$, each node in the tree will define a different state and the edges connecting two nodes define the actions. In this fashion, each state has two actions, which we denote $a_1$ and $a_2$, and taking either action leads deterministically to the corresponding child. Taking any action from states on the final level of the tree exits the MDP. The state $s_0$ is the root of the tree. We thus have defined the shared state and action space for MDPs supporting $\mathcal{D}$, and have also defined the shared and deterministic transition dynamics. So the MDPs are all deterministic by construction. Note that to verify \weakprox{} (a), the construction so far implies that the state-action space is shared and the transitions satisfy $\transvar = 0$, although we have not yet defined the reward structure so cannot say anything about $\rewvar$. \\

\noindent To complete the definition of $\mathcal{D}$, we must complete the definition of each individual MDP supporting $\mathcal{D}$ by defining a reward function for each MDP. We shall do so via the procedure described below. But before that, fix a state $s^*$ on level $H$, which is selected by sampling uniformly at random from the set of states on level $H$. Note that the location of $s^*$ will be kept hidden from the user, although we will define $\mathcal{D}$ with reference to a fixed $s^*$. \\

\noindent Note that there are $2^{\frac{H}{2} - 1} - 1$ states $s$ on level $\frac{H}{2}$ such that the subtree rooted at $s$ does not contain $s^*$. Also note that each subtree rooted at such an $s$ has $2^{\frac{H}{2} - 1}$ states on level $H$. If one were to ``view'' the final level of a binary tree on paper, as in Figure~\ref{fig:lower_det}, there is a natural ordering of the states on the final level from left to right. In each subtree rooted at such an $s$, consider the $i$th state in this ordering of the states on the final level of that subtree, and denote this $i$th state as $x_{i, s}$. Then define $\mathcal{S}_i = \cup_s x_{i, s}$. There are a total of $2^{\frac{H}{2} - 1}$ such sets $\mathcal{S}_i$. We will construct $2^{\frac{H}{2} - 1}$ MDPs total to support the distribution $\mathcal{D}$, one for each $\mathcal{S}_i$. Note that for $i \neq j$, we have $\mathcal{S}_i \cap \mathcal{S}_j = \varnothing$. Note also that $\mathcal{S}_i$ never contains a state which is also in the subtree rooted at level $\frac{H}{2}$ that contains $s^*$. \\

\noindent Define $\epsilon = \frac{1}{\frac{H}{2} - \log(H^k) - 1}$, and note that $\epsilon$ is $\Theta(\frac{1}{H})$. For each $\mathcal{S}_i$, we will define the MDP $M_i$ by defining $\Rew_{M_i}$ as follows. Fix an $\mathcal{S}_i$. For each $s \in \mathcal{S}_i$, let $s_p$ denote the unique ancestor of $s$ on level $\frac{H}{2} + \log(H^k) + 1$ of the tree. On the path connecting $s_p$ to $s$, assign each state-action pair a reward of $\epsilon$. So the total reward accumulated by following the path from the root through $s_p$ to $s$ and taking either action from $s$ is $\epsilon (\frac{H}{2} - \log(H^k) - 1) = 1$. Also, consider the unique ancestor $s_p^*$ of $s^*$ on level $\frac{H}{2} + \log(H^k) + 1$, and assign each state-action pair along the path $s_p^*$ to $s^*$ a reward of $\epsilon$. So the total reward accumulated by following the path from the root to $s^*$ and taking either action from $s^*$ is $\epsilon (\frac{H}{2} - \log(H^k) - 1) = 1$. Any other state-action pair in the tree is assigned zero reward. This completes the definition of $\Rew_{M_i}$, and hence the definition of $M_i$, except that we have not yet featurized the state-action space. \\

\noindent Perform this procedure for each of the $2^{\frac{H}{2} - 1}$ sets $\mathcal{S}_i$, to obtain a set which contains $2^{\frac{H}{2} - 1}$ such MDPs. Once we featurize the state-action space in the fashion described below, the definition of $\mathcal{D}$ is completed by assigning the uniform distribution to this set. The key behind this construction, is that for any MDP supporting $\mathcal{D}$, each subtree rooted at level $\frac{H}{2}$ contains a single path which provides the optimal unit value. But there is only one path providing unit value that is shared by each MDP, this is the path from the root to $s^*$. Note that for any $M_i$, all of the subtrees rooted at states on level $\frac{H}{2}$ are identical, with the exception of the subtree containing $s^*$. \\

Let us now discuss how to featurize the state-action space. Note that each MDP is defined on a common state-action space that has at most $2^H$ states. Sample vectors $\{ z_1, z_2 \dots z_{2^H} \}$ i.i.d from the spherical measure on the surface of the unit sphere that sits in $d-2$ dimensions. By the results of \citep{du20}, we can pick $d$ to be at most a polynomial of $H$, while ensuring that $\| z_i \|_2 = 1$ and $\vert z_i^T z_j \vert \leq \frac{1}{50}$ for all $i \neq j$. Take any ordering of the states $\{ s_i \}$ indexed by $i$, and assign the following features
$$
\phi(s_i, a_1) = [z_i, 1, 0] \text{ and } \phi(s_i, a_2) = [0, 1, 1].
$$
Rescaling the features to have unit norm is trivial, so we work with these since they make the computations more apparent. In the above, it appears that $\phi(s_i, a_2)$ is the same for each $i$, but this is without loss of generality since we can always add a dummy coordinate that makes them all unique. Crucially note that the features do \emph{not} depend on the reward structure of the MDP, since they are completely agnostic to the choice of rewards. Hence they do not leak any information about the rewards. \\

\noindent \textbf{Verifying \weakprox{} (a).} We have already checked above that the state-action space is shared and $\transvar = 0$. To see that $\rewvar$ is $\Theta(\frac{1}{H})$, simply note that any state in any MDP supporting $\mathcal{D}$ has reward either $0$ or $\epsilon$, and $\epsilon$ is $\Theta(\frac{1}{H})$. This verifies \weakprox{} (a). \\

\noindent \textbf{Verifying \weakprox{} (b).} Define $\pi^*$ to be the deterministic policy which prescribes the path leading from the root to $s^*$, and at states not along this path it prescribes an arbitrary action. It is then immediate from our above arguments that for any $M$, $\V{s_0}{M}{\pi^*} = \max_{\pi'} \V{s_0}{M}{\pi'} = 1$, so \weakprox{} (b) is satisfied with parameter $\alpha = 0$. \\

\noindent \textbf{Verifying SIO (a).} Fix any $M$. The key to verifying this is the observation that all subtrees rooted at level $H/2$ are identical (in terms of reward structure, not features), except for the one that contains $s^*$. In the first $H/2 - 1$ levels one can use an arbitrary policy. So now consider any level $h$ that is greater than or equal to $H/2$. Inside the subtree rooted at $H/2$ that contains $s^*$, there is a unique state $s$ on this level that is an ancestor of $s^*$. Let us denote $z$ to be the spherical measure random variable that was used to define the feature map for this state, so that $\phi(s, a_1) = [z, 1, 0] \text{ and } \phi(s, a_2) = [0, 1, 1]$. \\

\noindent Now observe that within the subtree rooted at $H/2$ that contains $s^*$, the only state from where one can achieve nonzero reward (on the level $h$ we are considering) is $s$. So the policy at the other states in this subtree does not matter. Similarly, in each of the other subtrees rooted at $H/2$, there is a unique state $s'$ which can yield nonzero reward (on the level $h$ we are considering). Furthermore, since these other subtrees are identical, the optimal policy within one subtree works for another. To construct our $\theta \equiv \theta^\star_h$ for this level $h$, we use this information to claim that there are four cases to consider.

\noindent First, consider the case when $\pi_M^*(s) = a_1$ and $\pi_M^*(s') = a_2$ for any $s'$ in another subtree which can yield nonzero reward. Let $\theta = [z, 0, 1/2]$. Then we have that
$$
\phi(s, a_1)^T \theta = [z, 1, 0]^T [z, 0, 1/2] = 1 + 0 + 0 = 1 \text{ and } \phi(s, a_2)^T \theta = [0, 1, 1]^T [z, 0, 1/2] = 0 + 0 + 1/2 = 1/2,
$$
which implies that we pick $a_1$ at $s$ if we follow the linear policy. But
$$
\phi(s', a_1)^T \theta = [z', 1, 0]^T [z, 0, 1/2] \leq \frac{1}{50} + 0 + 0 = \frac{1}{50} \text{ and } \phi(s', a_2)^T \theta = [0, 1, 1]^T [z, 0, 1/2] = 0 + 0 + 1/2 = 1/2,
$$
which implies that we pick $a_2$ at $s'$ if we follow the linear policy $\theta$. Note that this argument holds for \emph{any} $s'$ which can yield nonzero reward lying in \emph{any} of the subtrees that do not contain $s^*$, and also recall the optimal policy does not change when looking at different subtrees not containing $s^*$. \\

\noindent Second, consider the case when $\pi_M^*(s) = a_2$ and $\pi_M^*(s') = a_1$ for any $s'$ in another subtree which can yield nonzero reward. Let $\theta = -[z, 0, 1/2]$. Since we have simply negated the above case, the result holds. \\

\noindent Third, consider the case when $\pi_M^*(s) = a_1$ and $\pi_M^*(s') = a_1$ for any $s'$ in another subtree which can yield nonzero reward. Let $\theta = [0, 1, -1/2]$. Then the result follows immediately because $\phi(s'', a_1)^T\theta = 1$ but $\phi(s'', a_2)^T\theta = 1/2$ for every state $s''$ in the tree. \\

\noindent Fouth, consider the case when $\pi_M^*(s) = a_2$ and $\pi_M^*(s') = a_2$ for any $s'$ in another subtree which can yield nonzero reward. Let $\theta = -[0, 1, -1/2]$. Then the result follows immediately because we have simply negated the above case. \\

\noindent It remains to rescale the features and $\theta$ to ensure they all have unit norm. \\

\noindent \textbf{Verifying SIO (b).} Consider the following algorithm, which takes as input an MDP $M_i$ that is known to support $\mathcal{D}$, but it knows nothing else about $M_i$ (in particular it does not know which $\mathcal{S}_i$ was used to define $M_i$). It of course does not know anything about the location of $s^*$ other than the fact that $s^*$ was sampled uniformly at random to define $\mathcal{D}$. It has also not been allowed to query any other MDPs supporting $\mathcal{D}$. The algorithm begins by picking an arbitrary state $s'$ on level $\frac{H}{2}$. It then queries each state on level $\log(H^k) + 1$ of the subtree rooted at $s'$ (this is level $\frac{H}{2} + \log(H^k) + 1$ of the entire tree). There is a single state $s_p$ on this level which has an action providing reward $\epsilon$, and all other states will give reward zero. When it finds the state $s_p$ on this level providing reward $\epsilon$, it stores the path leading to this state $s_p$. It then queries each of the two child states of $s_p$ to identify which of these child states lies along the optimal path, which is doable since exactly one of these child states has an action which provides reward $\epsilon$. It greedily takes the action required to arrive at this child, and stores this action. It then repeats this greedy procedure from the child $\Theta(H)$ times until it reaches the final level of the tree. In this manner, if it was given MDP $M_i$ as input then it identifies a path from the root through $s'$ and $s_p$ (or perhaps $s^*_p$) to either $s^*$ or some state $s$ on level $H$ which satisfies $s \in \mathcal{S}_i$. Let $\pi$ be the deterministic policy which prescribes this path, and prescribes arbitrary actions for states not along this path, then it is clear $\V{s_0}{M_i}{\pi} = \max_{\pi'} \V{s_0}{M_i}{\pi'} = 1$. Note also that this same algorithm also immediately can be used to find $\pi$ satisfying $\V{a}{M_i}{\pi} = \max_{\pi'} \V{a}{M_i}{\pi'}$ for any initial state $a$. To see this, note that if $a$ is an ancestor of $s^*$ or some $s \in \mathcal{S}_i$, then the described algorithm will identify this and find an appropriate path. If $a$ is not an ancestor of either of these, then the value of this state is zero and the same algorithm can be used to certify this. The sample complexity of this algorithm is the number of queries it took at the beginning to identify $s_p$ in the subtree rooted at $s'$. Note that there are $2^{\log(H^k) + 1}$ states on level $\log(H^k) + 1$ of the subtree rooted at $s'$, so the sample complexity of this algorithm is $\mathcal{O}(H^k)$. To convert the policy $\pi$ found by the algorithm into a linear policy $\overline{\theta} = \{ \theta_h \}_{h=0}^{H-1}$, simply note that we can set $\theta_h = [0, 1, - \frac{1}{2}]$ if $\pi$ recommends $a_1$ while following its path at timestep $h$, and set $\theta_h = [0, -1, \frac{1}{2}]$ if $\pi$ recommends $a_2$ while following its path at timestep $h$. Then rescale $\theta_h$ to ensure it has unit norm. This verifies SIO (b) with parameter $\beta = 0$. \\

\noindent With this construction of $\mathcal{D}$ in hand, we return to the proof of Theorem~\ref{thm:avg_per_det_lower_bound}, during which we shall also prove the claim we made above about how $s^*$ cannot be identified in a polynomial number of samples. A basic computation reveals that any path through the tree which does not end in $s^*$, has value (in expectation over $\mathcal{D}$ w.r.t. $s_0$) at most $\frac{2}{H}$. However, the path through the tree which ends in $s^*$ has value (in expectation over $\mathcal{D}$ w.r.t. $s_0$) precisely one. The optimal policy (in terms of value in expectation over $\mathcal{D}$ w.r.t. $s_0$) is hence clearly the deterministic policy which prescribes following the path through the tree that ends in $s^*$. Hence, any algorithm which can find a policy $\pi$ satisfying $\E_{M \sim \mathcal{D}} [\V{s_0}{M}{\pi}] \geq \max_{\pi'} \E_{M \sim \mathcal{D}} [\V{s_0}{M}{\pi'}] - \frac{1}{4}$ must be able to identify $s^*$ with non-trivial probability. So to show that finding such a $\pi$ requires $\Omega(2^H)$ queries, it is sufficient to show that identifying $s^*$ requires $\Omega(2^H)$ queries. \\

\noindent This is immediate from the nature of our construction. Simply observe that any algorithm must query in the subtree rooted at level $\frac{H}{2}$ that contains $s^*$ at least once in order to identify $s^*$. However, for any MDP there are $2^{\frac{H}{2} - 1}$ subtrees on level $\frac{H}{2}$ and all but one of them are identical. Of course, identifying the correct subtree with non-trivial probability requires $\Omega(2^H)$ total queries. Note that we crucially used here the fact that the features do \emph{not} depend on the reward structure of the MDP, and hence do not leak any information about the rewards. Note also that our argument holds when the agent has access to a generative model, since it can transition to any state to query it. Moreover, the difficulty here comes from querying states, and not from sampling MDPs. So the result holds under SQM. Finally, we note that the policy prescribing the path through $s^*$ can easily be expressed as a linear policy. This completes the proof.

\subsection{Proof Of Corollary~\ref{corollary:meta_rl_det}}
\label{app:lower_corr_meta_det}
We use the same $\mathcal{D}$ that was constructed in the proof of Theorem~\ref{thm:avg_per_det_lower_bound}. Before proceeding with the proof of Corollary~\ref{corollary:meta_rl_det}, we note the following key fact. If we sample $M$ from $\mathcal{D}$, but do not query any MDP supporting $\mathcal{D}$ beforehand, then any algorithm that can find (possibly nonlinear) $\pi$ satisfying $\V{s_0}{M}{\pi} \geq \max_{\pi'} \V{s_0}{M}{\pi'} - \nicefrac{1}{4}$, with probability at least $\nicefrac{1}{2}$ over the selection of $M$, requires $\Omega(H^k)$ query cost (under QM) on $M$. This fact demonstrates that the algorithm described in SIO (b) is minimax optimal, since \emph{any} procedure will need the same complexity to solve an MDP sampled from $\mathcal{D}$, assuming it has not already queried other MDPs beforehand. In particular, any algorithm which can solve $\Mtest$ at test time in $o(H^k)$ queries must rely on querying during training time. \\

To prove this fact, note that any algorithm (that is optimal to within the constant $\frac{1}{4}$) must discover a path that has $\Omega(H)$ state-action pairs intersecting with a path that leads either to $s^*$ or to a state $s \in \mathcal{S}_i$. Of course, discovering such a path is equivalent to identifying $s^*$ or the $s_p$ corresponding to $s \in \mathcal{S}_i$. Assume for now the claim that we cannot identify $s^*$ with a polynomial number of samples. Then we need only show that identifying an $s_p$ corresponding to $s \in \mathcal{S}_i$ requires $\Omega(H^k)$ queries. But this is immediate, since all we know is that $M_i$ was sampled from $\mathcal{D}$, and each subtree (that doesn't contain $s^*$) rooted at level $\frac{H}{2}$ is identical. So without loss of generality take any subtree (not containing $s^*$) rooted at level $\frac{H}{2}$, then a priori our prior for the location of $s_p$ in that subtree is exactly the uniform distribution over states on level $\log(H^k) + 1$ of that subtree. So we must query at least half of the $2^{\log(H^k) + 1} = \Omega(H^k)$ states on level $\log(H^k) + 1$ of that subtree to identify $s_p$ with non-trivial probability. Conditioned on our claim that it is not easy to identify $s^*$, the claimed fact. Note that this argument uses the fact that we can directly query states on level $\log(H^k) + 1$ of that subtree, and so holds under a generative model. Note that we crucially used here the fact that the features do \emph{not} depend on the reward structure of the MDP, and hence do not leak any information about the rewards. \\

We now return to the proof of Corollary~\ref{corollary:meta_rl_det}. For any $M_i$, there are two types of paths that are optimal, the first is through $s^*$ and the others are through the states lying in $\mathcal{S}_i$. Note that since $\Mtest \sim \mathcal{D}$ at test time, the location of the optimal paths in $\Mtest$ that do not intersect $s^*$ are sampled uniformly at random. A nearly identical argument to the one used in the proof of Theorem~\ref{thm:avg_per_det_lower_bound}, and the proof of the preceding fact, shows that identifying any of these optimal paths that are sampled uniformly at random requires $\Omega(H^k)$ queries. The only difference is that we condition on the event that $\Mtest$ is not the same as any of the MDPs queried during training, which occurs with high probability when we are only allowed a polynomial number of queries during training time. Note that even after conditioning on this event, finding an optimal path (which does not go through $s^*$) of $\Mtest$ requires $\Omega(H^k)$ queries. This is simply because conditioning on the aforementioned event only changes the probability of the location of such a path by a value which is exponentially small in $H$, assuming again that we used polynomial in $H$ number of queries during training. So we cannot perform any inference to reduce the size of the set of feasible locations of an optimal path (which does not go through $s^*$). Then we can essentially repeat the same argument used in the proof of Theorem~\ref{thm:avg_per_det_lower_bound}. Hence, if an algorithm hopes to solve $\Mtest$ at test time in $o(H^k)$ queries, then during training time it must narrow the possible locations of $s^*$ to a set whose cardinality is polynomial in $H$. But by the proof of Theorem~\ref{thm:avg_per_det_lower_bound}, this would require $\Omega(2^H)$ queries.

\subsection{Proof Of Corollary~\ref{corollary:avg_per_stoch}}
\label{app:lower_corr_avg_per_stoch}

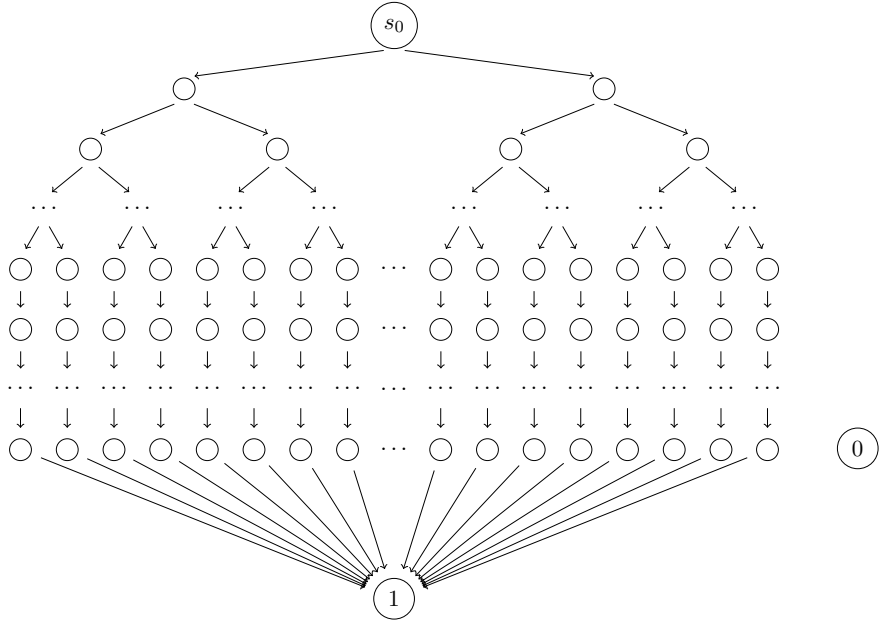
\begin{figure}[!tbp]
\centering
\scalebox{0.8}{
\begin{tikzpicture}[->,shorten <=5pt,shorten >=5pt,level distance=1cm]
\tikzset{}
\Tree
[.\node[draw,circle] {\(s_{0}\)};
    [.\node[draw,circle] {};
        [.\node[draw,circle] {};
            [.\(\ldots\)
                [.\node[draw,circle] {};
                    [.\node[draw,circle] {};
                        [.\(\ldots\)
                            [.\node[draw,circle] (t1) {}; ]
                        ]
                    ]
                ]
                [.\node[draw,circle] {};
                    [.\node[draw,circle] {};
                        [.\(\ldots\)
                            [.\node[draw,circle] (t2) {}; ]
                        ]
                    ]
                ]
            ]
            [.\(\ldots\)
                [.\node[draw,circle] {};
                    [.\node[draw,circle] {};
                        [.\(\ldots\)
                            [.\node[draw,circle] (t3) {}; ]
                        ]
                    ]
                ]
                [.\node[draw,circle] {};
                    [.\node[draw,circle] {};
                        [.\(\ldots\)
                            [.\node[draw,circle] (t4) {}; ]
                        ]
                    ]
                ]
            ]
        ]
        [.\node[draw,circle] {};
            [.\(\ldots\)
                [.\node[draw,circle] {};
                    [.\node[draw,circle] {};
                        [.\(\ldots\)
                            [.\node[draw,circle] (t5) {}; ]
                        ]
                    ]
                ]
                [.\node[draw,circle] {};
                    [.\node[draw,circle] {};
                        [.\(\ldots\)
                            [.\node[draw,circle] (t6) {}; ]
                        ]
                    ]
                ]
            ]
            [.\(\ldots\)
                [.\node[draw,circle] {};
                    [.\node[draw,circle] {};
                        [.\(\ldots\)
                            [.\node[draw,circle] (t7) {}; ]
                        ]
                    ]
                ]
                [.\node[draw,circle] {};
                    [.\node[draw,circle] {};
                        [.\(\ldots\)
                            [.\node[draw,circle] (t8) {}; ]
                        ]
                    ]
                ]
            ]
        ]
    ]
\edge[transparent] node{}; [
    .{} \edge[transparent] node{};
        [.{} \edge[transparent] node{};
            [.{} \edge[transparent] node{};
                [.{\(\ldots\)} \edge[transparent] node{};
                    [.{\(\ldots\)} \edge[transparent] node{};
                        [.{\(\ldots\)} \edge[transparent] node{};
                            [.\node[draw=none] (mid) {\(\ldots\)}; ]
                        ]
                    ]
                ]
            ]
        ]
    ]
\edge[] node{};
    [.\node[draw,circle] {};
        [.\node[draw,circle] {};
            [.\(\ldots\)
                [.\node[draw,circle] {};
                    [.\node[draw,circle] {};
                        [.\(\ldots\)
                            [.\node[draw,circle] (t9) {}; ]
                        ]
                    ]
                ]
                [.\node[draw,circle] {};
                    [.\node[draw,circle] {};
                        [.\(\ldots\)
                            [.\node[draw,circle] (t10) {}; ]
                        ]
                    ]
                ]
            ]
            [.\(\ldots\)
                [.\node[draw,circle] {};
                    [.\node[draw,circle] {};
                        [.\(\ldots\)
                            [.\node[draw,circle] (t11) {}; ]
                        ]
                    ]
                ]
                [.\node[draw,circle] {};
                    [.\node[draw,circle] {};
                        [.\(\ldots\)
                            [.\node[draw,circle] (t12) {}; ]
                        ]
                    ]
                ]
            ]
        ]
        [.\node[draw,circle] {};
            [.\(\ldots\)
                [.\node[draw,circle] {};
                    [.\node[draw,circle] {};
                        [.\(\ldots\)
                            [.\node[draw,circle] (t13) {}; ]
                        ]
                    ]
                ]
                [.\node[draw,circle] {};
                    [.\node[draw,circle] {};
                        [.\(\ldots\)
                            [.\node[draw,circle] (t14) {}; ]
                        ]
                    ]
                ]
            ]
            [.\(\ldots\)
                [.\node[draw,circle] {};
                    [.\node[draw,circle] {};
                        [.\(\ldots\)
                            [.\node[draw,circle] (t15) {}; ]
                        ]
                    ]
                ]
                [.\node[draw,circle] {};
                    [.\node[draw,circle] {};
                        [.\(\ldots\)
                            [.\node[draw,circle] (t16) {}; ]
                        ]
                    ]
                ]
            ]
        ]
    ]
]
\node[draw,circle,below= 2cm of mid] (end) {\(1\)};
\foreach \num in {1,...,16}{
    \draw (t\num) -> (end);
}
\node[draw, circle, right=7cm of mid] (dummy) {\(0\)};
\end{tikzpicture}
}
\caption{An illustration of the generic structure used to define the MDPs that support the $\mathcal{D}$ constructed in the proof of Corollary~\ref{corollary:avg_per_stoch}. Observe that the first half of the structure is a tree, while the second half comprises of linear sequences of states.}
\label{fig:lower_stoch}
\end{figure}

\textbf{Construction Details.} The $\mathcal{D}$ that we construct here is very similar in spirit to the $\mathcal{D}$ that was constructed in the proof of Theorem~\ref{thm:avg_per_det_lower_bound}. $\mathcal{D}$ will be supported by MDPs whose shared state and action spaces all share the same generic structure. We depict this generic structure in Figure~\ref{fig:lower_stoch}. For the first $\frac{H}{2}$ levels, the structure is defined by a binary tree, where the nodes in the tree represent states. Each state in the tree has two actions. For the next $\frac{H}{2}$ levels, there are numerous linear sequences of states of length $\frac{H}{2}$. Each such sequence emanates from a corresponding state that is a leaf of the tree. These sequences all end in a common state, which we denote $\circled{1}$. Each state in the sequence has a single action. The state $s_0$ is the root of the tree. There is also a state $\circled{0}$, which a priori is disconnected from the remainder of the structure. This implies that there is a shared state-action space. For each MDP $M$ supporting $\mathcal{D}$, we will have that $\Rew_M$ maps $\circled{1}$ to one, and maps all other states including $\circled{0}$ to zero. Taking any action from $\circled{0}$ or $\circled{1}$ always exits the MDP for any MDP supporting $\mathcal{D}$. A priori, the transition function in this generic structure is deterministic. And it is ``natural" in the sense that taking either action from a state in the tree leads to the corresponding child, and taking the action when in the linear sequence of states leads to the next state in the sequence. Of course, we will modify this ``a priori natural" transition function in different ways for each MDP. \\

\noindent Let $\mathcal{F}_h$ denote the set of states at level $h$ of this generic structure that we have outlined above. Before constructing $\mathcal{D}$, first select a state $s^*$ uniformly at random from the $2^{\nicefrac{H}{2} - 1}$ states in $\mathcal{F}_{\nicefrac{H}{2}}$. As in the proof of Theorem~\ref{thm:avg_per_det_lower_bound}, $s^*$ is hidden from the user. \\

\noindent Note that there are $2^{\frac{H}{4} - 1} - 1$ states $s$ on level $\frac{H}{4}$ such that the subtree rooted at $s$ does not contain $s^*$. Also note that each subtree rooted at such an $s$ has $2^{\frac{H}{4} - 1}$ states on level $\frac{H}{2}$ of the entire structure. If one were to ``view" the final level of a binary tree on paper, there is a natural ordering of the states on the final level from left to right. In each subtree rooted at such an $s$ such that subtree does not contain $s^*$, consider the $i$th state in this ordering of the states on the final level of that subtree (which is the level $\frac{H}{2}$ of the entire structure). Denote this $i$th state as $x_{i, s}$. Then define $\mathcal{S}_i = \cup_s x_{i, s}$. There are a total of $2^{\frac{H}{4} - 1}$ such sets $\mathcal{S}_i$. We will construct $2^{\frac{H}{4} - 1}$ MDPs total to support the distribution $\mathcal{D}$, one for each $\mathcal{S}_i$. Note that for $i \neq j$, we have $\mathcal{S}_i \cap \mathcal{S}_j = \varnothing$. Note also that $\mathcal{S}_i$ never contains a state which is also in the subtree rooted at level $\frac{H}{4}$ that contains $s^*$. \\

\noindent To define each MDP $M$ that supports $\mathcal{D}$, it is sufficient to define the transition function $\Trans_M$ for $M$, since we have already defined the shared state-action space and the shared reward function. Similar to the proof of Theorem~\ref{thm:avg_per_det_lower_bound}, for each $\mathcal{S}_i$ construct an MDP $M_i$ as follows:
\begin{enumerate}
\item Consider the linear portion of $M_i$ that lies below any state $s'$ satisfying $s' \notin \mathcal{S}_i$ and $s' \neq s^*$. Recall that a priori, the generic structure for any MDP was deterministic, which implies when you are at a state in this linear portion then taking the action deterministically leads to the next state in the sequence. We now modify this so that for any state-action pair along the linear sequence below $s'$, taking the action takes you to the child with probability $1 - \nicefrac{10}{H}$, and makes you jump out to state $\circled{0}$ with probability $\nicefrac{10}{H}$. Do this for all $s'$ satisfying $s' \notin \mathcal{S}_i$ and $s' \neq s^*$. Leave the linear sequence of states below $s^*$ and any $s \in \mathcal{S}_i$ unchanged, so that the transitions in these two linear sequences remain deterministic.
\item For each $s \in \mathcal{S}_i$, there is a unique path connecting state $s$ to its unique ancestor on level $\frac{H}{4}$. This path defines a sequence of state-action pairs that leads one from level $\frac{H}{4}$ to $s$. Again recall that a priori, the generic structure for any MDP was deterministic, which implies that taking one of these actions while in the tree deterministically leads you to the corresponding child. Modify the transitions for these state-action pairs along this path of length $\frac{H}{4}$, so that when you take an action, then with probability $1 - \nicefrac{1}{H^{k-1}}$ the action leads to the child, and with probability $\nicefrac{1}{H^{k-1}}$ it takes you directly to state $\circled{1}$. Do this for each $s \in \mathcal{S}_i$. Similarly, consider the path between $s^*$ and its unique ancestor on level $\frac{H}{4}$. Modify the transitions for the state-action pairs along this path so that when you take an action, then with probability $1 - \nicefrac{1}{H^{k-1}}$ the action leads to the child, and with probability $\nicefrac{1}{H^{k-1}}$ it takes you directly to state $\circled{1}$.
\end{enumerate}
We have thus defined $\Trans_{M_i}$ for each $M_i$, where each $M_i$ is identified by a particular $\mathcal{S}_i$. This defines a set containing a total of $2^{\frac{H}{4} - 1}$ different MDPs $M_i$. To complete the definition of $\mathcal{D}$, simply consider the uniform distribution over the set of $M_i$ that we have created. We use a featurization that is identical to the one used in Theorem~\ref{thm:avg_per_det_lower_bound} for the binary tree portion of our construction here, and arbtirary features for the linear portion since there is only one action to take here. \\

\noindent \textbf{Verifying \weakprox{} (a).} We have already established the state-action space for each MDP is shared. Note that by definition each MDP shares a reward function, since $\circled{1}$ always provides unit reward and each other state provides zero reward, regardless of the MDP. So we have $\rewvar = 0$. Then observe that when we modified the transitions at states, we changed them by making them lead to states $\circled{0}$ or $\circled{1}$ with probability at most $\max \{ \frac{1}{H^k}, \frac{10}{H} \}$. It remains to note the $\ell_1$ characterization of the total variation distance and that $k \geq 3$, implying that $\transvar = \Theta(\frac{1}{H})$. This verifies \weakprox{} (a). \\

\noindent \textbf{Verifying \weakprox{} (b).} Observe that following the unique path leading from the root through state $s^*$ till $\circled{1}$ always provides value $1$, regardless of the MDP. This is because if we take the actions corresponding to this path, then either we hop out to $\circled{1}$ while we are in the tree, or we reach the linear sequence from where we deterministically arrive at $\circled{1}$. Hence this path is always optimal. In turn, the policy which prescribes this path, while doing something arbitrary at states not along this path, always provides optimal value. So Weak Proximity (b) is satisfied with parameter $\alpha = 0$. \\

\noindent \textbf{Verifying SIO (a).} The proof of this is identical to the one in Theorem~\ref{thm:avg_per_det_lower_bound}. \\

\noindent \textbf{Verifying SIO (b).} To verify this, sample $M_i \sim \mathcal{D}$ and consider the following greedy algorithm. Note that $s^*$ is not revealed a priori, nor do we know the locations of the states in $\mathcal{S}_i$, and the only information that the user has is that this MDP supports $\mathcal{D}$. Otherwise this problem trivially does not require any samples to solve. Start at any arbitrary state on level $\frac{H}{4}$ and sample the left action $\mathcal{O}(H^{k-1})$ times and the right action $\mathcal{O}(H^{k-1})$ times. With high probability, one of these actions will lead to the state $\circled{1}$ at least once. And of course with unit probability the other action will not lead to $\circled{1}$. Simply take the action that has led to $\circled{1}$ at least once, and repeat this procedure at the next state. Repeating this procedure until you reach level $\nicefrac{H}{2}$ and union bounding guarantees that with high probability we find a path that leads to either $s^*$ or a state $s \in \mathcal{S}_i$. From here on, one can deterministically follow the linear sequence of states to arrive at $\circled{1}$. The policy that prescribes this path then provides the optimal (unit) value for that MDP, so $\beta = 0$. The total sample complexity of this method is $\Theta(H^{k-1} \times H) = \Theta(H^k)$. Again, note that this argument holds under a generative model as defined in SQM, since we can plug in whatever state we want, and receive as feedback from the generative model the next state sampled from the transition process. Note also that in similar fashion to Theorem~\ref{thm:avg_per_det_lower_bound}, this same algorithm can be used to identify a policy achieving optimal value with respect to any initial state in the tree. To convert the policy found by the algorithm into a linear policy, we use the same technique as in Theorem~\ref{thm:avg_per_det_lower_bound}. \\

\noindent We now complete the proof of Corollary~\ref{corollary:avg_per_stoch}. A basic computation shows that for $k \geq 3$ and sufficiently large $H$, the value (in expectation over $\mathcal{D}$) of any path is at most $\frac{1}{10}$. But the value of the path (in expectation over $\mathcal{D}$) through $s^*$ till $\circled{1}$ is one. Hence any algorithm which can find $\pi$ satisfying $\E_{M \sim \mathcal{D}} [\V{s_0}{M}{\pi}] \geq \max_{\pi'} \E_{M \sim \mathcal{D}} [\V{s_0}{M}{\pi'}] - \frac{1}{4}$ must be able to identify $s^*$ with non-trivial probability. So to show that finding such a $\pi$ requires $\Omega(2^H)$ queries, it is sufficient to show that identifying $s^*$ requires $\Omega(2^H)$ queries. This is done via identical arguments to the ones used to prove Theorem~\ref{thm:avg_per_det_lower_bound}.

\subsection{Proof Of Corollary~\ref{corollary:meta_rl_stoch}}
\label{app:lower_corr_meta_stoch}
The proof of this corollary is basically identical to the proof of Corollary~\ref{corollary:meta_rl_det}. Briefly, if an algorithm can solve $\Mtest$ with a number of queries at test time that is strictly fewer than $\Omega(H^k)$, then at training time it must have narrowed the possible locations of $s^*$ to a set whose cardinality is polynomial in $H$. By the proofs of Theorem~\ref{thm:avg_per_det_lower_bound} and Corollary~\ref{corollary:avg_per_stoch}, this requires $\Omega(2^H)$ queries during training time.

\section{Upper Bound Proof Details}
\label{app:upper}
In this section, we will provide a formal proof of Theorem~\ref{thm:upper_bound}. For ease in presentation, we shall assume throughout that the action space $\ActSet$ for each MDP contains two actions, which we denote $a_1$ and $a_2$. It is easy to extend the proofs to the case when there are many actions. We will often use the notation $s_1$ and $s_2$ to denote the child states of a state $s$ when taking actions $a_1$ and $a_2$ respectively. We shall also use $\pi^*_M$ to denote an optimal policy for MDP $M$. Before proving Theorem~\ref{thm:upper_bound}, we shall state two helpful lemmas. Recall the definition of $\pi^*$ from \strprox{} (b).

\begin{lemma}
\label{lemma:upper_bound_helper_one}
Consider any $\mathcal{D}$ satisfying WIO with $\beta \geq 0$ and \strprox{} with $\transvar = 0$ and any $\polsub, \rewvar \geq 0$, such that the MDPs supporting $\mathcal{D}$ are deterministic. Run Algorithm~\ref{alg:main} with $n \geq 1$ samples, and assume at timestep $t$ we are at state $s_t$ such that $\pi^*(s_t) = a_1$. We are guaranteed that the event
$$
\frac{1}{n} \sum_{i=1}^n Q_{i,a_1} \geq \frac{1}{n} \sum_{i=1}^n Q_{i,a_2} - \polsub - \solvesub
$$
occurs almost surely. The symmetric statement for $\pi^*(s_t) = a_2$ is also true.
\end{lemma}

\begin{lemma}
\label{lemma:upper_bound_helper_two}
Consider any $\mathcal{D}$ satisfying WIO with $\beta \geq 0$ and \strprox{} with $\transvar = 0$ and any $\polsub, \rewvar \geq 0$, such that the MDPs supporting $\mathcal{D}$ are deterministic. Run Algorithm~\ref{alg:main} with $n = \frac{H^2}{\epsilon^2} \log \left( \frac{2 H \vert \ActSet \vert}{\delta} \right)$ samples, and assume at timestep $t$ we are at state $s \equiv s_t$. Then the event
\begin{equation*}
\begin{split}
&\left \vert \frac{1}{n} \sum_{i=1}^n Q_{i, a_1} - \E_{M \sim \mathcal{D}} \left[ \Rew_M(s, a_1) + \V{s_1}{M}{\pi^*_M} \right] \right \vert \leq \solvesub + \frac{\epsilon}{2H} \\
\text{ and } &\left \vert \frac{1}{n} \sum_{i=1}^n Q_{i, a_2} - \E_{M \sim \mathcal{D}} \left[ \Rew_M(s, a_2) + \V{s_2}{M}{\pi^*_M} \right] \right \vert \leq \solvesub + \frac{\epsilon}{2H}
\end{split}
\end{equation*}
occurs with probability at least $1 - \frac{\delta}{H}$.
\end{lemma}

\noindent The proofs of Lemmas~\ref{lemma:upper_bound_helper_one} and~\ref{lemma:upper_bound_helper_two} can be found in Appendix~\ref{app:upper_lem1} and Appendix~\ref{app:upper_lem2} respectively. With these lemmas in hand, we now turn to the proof of Theorem~\ref{thm:upper_bound}.

\subsection{Proof Of Theorem~\ref{thm:upper_bound}}
\label{app:upper_thm_proof}
\noindent First observe that by \strprox{} (b), we have
$$
\E_{M \sim \mathcal{D}} \left[ \V{s_0}{M}{\pi^*} \right] \geq \E_{M \sim \mathcal{D}} \left[ \max_{\pi_M} \V{s_0}{M}{\pi_M} \right] - \polsub \geq \max_{\pi'} \E_{M \sim \mathcal{D}} \left[ \V{s_0}{M}{\pi'} \right] - \polsub.
$$
Hence to prove the theorem, it is sufficient to prove that
$$
\E_{M \sim \mathcal{D}} \left[ \V{s_0}{M}{\pi} \right] \geq \E_{M \sim \mathcal{D}} \left[ \V{s_0}{M}{\pi^*} \right] - \epsilon - 2 \polsub H - 3 \solvesub H,
$$
and we shall devote the remainder of the proof to this. \\

\noindent Recall that the policy constructed by Algorithm~\ref{alg:main} is represented as a vector of length $H$, which stores an action for each timestep. We use the terminology ``algorithm recommends an action at a timestep'' to mean that at that timestep, the algorithm stores that action in the policy that it is constructing. Our proof rests on a key claim, which is that while following Algorithm~\ref{alg:main}, at each timestep the algorithm recommends an action whose suboptimality (in expectation over all MDPs) relative to the other action is at most $\frac{\epsilon}{H} + 2\polsub + 3\solvesub$. Concretely, assume the algorithm recommends an action $a$ at state $s$ which transports us to $s'$. We claim that with high probability at least $1 - \frac{\delta}{H}$,
\begin{equation}
\label{eqn:upper_bound_claim}
\E_{M \sim \mathcal{D}} \left[ \Rew_M(s, a) + \V{s'}{M}{\pi^*} \right] \geq \E_{M \sim \mathcal{D}} \left[ \V{s}{M}{\pi^*} \right] - \frac{\epsilon}{H} - 2 \polsub - 3 \solvesub.
\end{equation}
First, we argue why this claim in Eq.~\eqref{eqn:upper_bound_claim} is sufficient to prove the theorem. Assume this claim to be true, and denote $s_h$ to be the state achieved by the algorithm after $h$ timesteps and $a_h$ to be the action recommended by the algorithm at timestep $h$. Then applying a union bound over $H$ timesteps, with high probability at least $1 - \delta$ we are guaranteed that
\begin{align*}
\E_{M \sim \mathcal{D}} \left[ \V{s_{H}}{M}{\pi^*} + \sum_{h=0}^{H-1} \Rew_M(s_h, a_h) \right] &= \E_{M \sim \mathcal{D}} \left[ \Rew_M(s_{H-1}, a_{H-1}) + \V{s_{H}}{M}{\pi^*} \right] + \E_{M \sim \mathcal{D}} \left[ \sum_{h=0}^{H-2} \Rew_M(s_h, a_h) \right] \\
&\geq \E_{M \sim \mathcal{D}} \left[ \V{s_{H-1}}{M}{\pi^*} + \sum_{h=0}^{H-2} \Rew_M(s_h, a_h) \right] - \frac{\epsilon}{H} - 2 \polsub - 3 \solvesub \\
&\geq \E_{M \sim \mathcal{D}} \left[ \V{s_{H-2}}{M}{\pi^*} + \sum_{h=0}^{H-3} \Rew_M(s_h, a_h) \right] - \frac{2\epsilon}{H} - 4\polsub - 6 \solvesub \\
&\geq \E_{M \sim \mathcal{D}} \left[ \V{s_0}{M}{\pi^*} \right] - \frac{\epsilon H}{H} - 2\polsub H - 3 \solvesub H \\
&= \E_{M \sim \mathcal{D}} \left[ \V{s_0}{M}{\pi^*} \right] - \epsilon - 2\polsub H - 3 \solvesub H.
\end{align*}
By assumption, the MDPs supporting $\mathcal{D}$ have shared deterministic transitions and a common state-action space $\StateSet \times \ActSet$, and hence the above calculations remain valid. So we have found a sequence of actions $\{ a_h \}_{h=0}^{H-1}$, which defines a path through the $\StateSet \times \ActSet$ and enables us to arrive at state $s_{H} \in \StateSet$ with the above property. Of course $\V{s_{H}}{M}{\pi^*}$ is just trivially zero, since the planning horizon is $H$. So the path $\{ a_h \}_{h=0}^{H-1}$ we have found, which defines a deterministic policy denoted by $\pi$, satisfies
\begin{align*}
\E_{M \sim \mathcal{D}} \left[ \V{s_0}{M}{\pi} \right] &= \E_{M \sim \mathcal{D}} \left[ \sum_{h=0}^{H-1} \Rew_M(s_h, a_h) \right] \\
&= \E_{M \sim \mathcal{D}} \left[ \V{s_{H}}{M}{\pi^*} + \sum_{h=0}^{H-1} \Rew_M(s_h, a_h) \right] \\
&\geq \E_{M \sim \mathcal{D}} \left[ \V{s_0}{M}{\pi^*} \right] - \epsilon - 2\polsub H - 3 \solvesub H,
\end{align*}
which exactly proves the theorem. \\

\noindent Hence it is sufficient to prove the claim in Eq.~\eqref{eqn:upper_bound_claim}, and we shall devote the remainder to proving this claim. Assume that while running the algorithm we are at some state $s$. Recall the notation that $s_1$ and $s_2$ are the child states of $s$ when taking actions $a_1$ and $a_2$ respectively, and recall $\pi^*_{M}$ denotes an optimal policy for MDP $M$. By the result of Lemma~\ref{lemma:upper_bound_helper_two}, we have with probability at least $1 - \frac{\delta}{H}$ that
\begin{equation}
\label{eqn:upper_bound_hoeffding}
\begin{split}
&\left \vert \frac{1}{n} \sum_{i=1}^n Q_{i, a_1} - \E_{M \sim \mathcal{D}} \left[ \Rew_M(s, a_1) + \V{s_1}{M}{\pi^*_M} \right] \right \vert \leq \solvesub + \frac{\epsilon}{2H} \\
\text{ and } &\left \vert \frac{1}{n} \sum_{i=1}^n Q_{i, a_2} - \E_{M \sim \mathcal{D}} \left[ \Rew_M(s, a_2) + \V{s_2}{M}{\pi^*_M} \right] \right \vert \leq \solvesub + \frac{\epsilon}{2H}.
\end{split}
\end{equation}
We condition on this event to verify the claim in Eq.~\eqref{eqn:upper_bound_claim}. \\

\noindent Now assume WLOG that $\pi^*(s) = a_1$, since the case when $\pi^*(s) = a_2$ is entirely symmetric. By the principle of Bellman optimality, this ensures that for any MDP $M$ we have $\Rew_M(s, a_1) + \V{s_1}{M}{\pi^*} = \V{s}{M}{\pi^*}$. Furthermore by the result of Lemma~\ref{lemma:upper_bound_helper_one},
\begin{equation}
\label{eqn:upper_bound_Q1_lower_bound}
\frac{1}{n} \sum_{i=1}^n Q_{i,a_1} \geq \frac{1}{n} \sum_{i=1}^n Q_{i,a_2} - \polsub - \solvesub.
\end{equation}

\noindent We now consider two cases. For the first case, assume $\frac{1}{n} \sum_{i=1}^n Q_{i,a_1} > \frac{1}{n} \sum_{i=1}^n Q_{i,a_2}$. Then the algorithm recommends action $a_1$ and transports us to state $s_1$. Recall again by our WLOG assumption that for any MDP $M$, we have $\Rew_M(s, a_1) + \V{s_1}{M}{\pi^*} = \V{s}{M}{\pi^*}$. So we are guaranteed that
$$
\E \left[ \Rew_M(s, a_1) + \V{s_1}{M}{\pi^*} \right] = \E \left[ \V{s}{M}{\pi^*} \right],
$$
and so the claim in Eq.~\eqref{eqn:upper_bound_claim} is trivially shown to be true in this case. \\

\noindent For the second case, assume $\frac{1}{n} \sum_{i=1}^n Q_{i,a_1} \leq \frac{1}{n} \sum_{i=1}^n Q_{i,a_2}$. Then the algorithm will recommend action $a_2$ and transport us to state $s_2$. But note that by the bound Eq.~\eqref{eqn:upper_bound_Q1_lower_bound}, we have
$$
\left \vert \frac{1}{n} \sum_{i=1}^n Q_{i,a_1} - \frac{1}{n} \sum_{i=1}^n Q_{i,a_2} \right \vert \leq \polsub + \solvesub.
$$
Combining this with the bound Eq.~\eqref{eqn:upper_bound_hoeffding}, and the triangle inequality, we have
$$
\left \vert \E_{M \sim \mathcal{D}} \left[ \Rew_M(s, a_1) + \V{s_1}{M}{\pi^*_M} \right] - \E_{M \sim \mathcal{D}} \left[ \Rew_M(s, a_2) + \V{s_2}{M}{\pi^*_M} \right] \right \vert \leq \polsub + 3\solvesub + \frac{\epsilon}{H}.
$$
Hence we have
\begin{align}
\label{eqn:upper_bound_last_eqn}
\begin{split}
\E_{M \sim \mathcal{D}} \left[ \Rew_M(s, a_2) + \V{s_2}{M}{\pi^*_M} \right] &\geq \E_{M \sim \mathcal{D}} \left[ \Rew_M(s, a_1) + \V{s_1}{M}{\pi^*_M} \right] - \polsub - 3 \solvesub - \frac{\epsilon}{H} \\
&\geq \E_{M \sim \mathcal{D}} \left[ \Rew_M(s, a_1) + \V{s_1}{M}{\pi^*} \right] - \polsub - 3 \solvesub - \frac{\epsilon}{H} \\
&= \E_{M \sim \mathcal{D}} \left[ \V{s}{M}{\pi^*} \right] - \polsub - 3 \solvesub - \frac{\epsilon}{H},
\end{split}
\end{align}
where the second inequality follows from the optimality of $\pi^*_M$ for $M$ and the equality follows from our WLOG assumption and the principle of Bellman optimality. Of course \strprox{} (b) also guarantees that for any MDP $M$ we have $\V{s_2}{M}{\pi^*} \geq \V{s_2}{M}{\pi^*_M} - \polsub$. We use this to upper bound the LHS of Eq.~\eqref{eqn:upper_bound_last_eqn} and obtain
$$
\E_{M \sim \mathcal{D}} \left[ \Rew_M(s, a_2) + \V{s_2}{M}{\pi^*} \right] \geq \E_{M \sim \mathcal{D}} \left[ \Rew_M(s, a_2) + \V{s_2}{M}{\pi^*_M} \right] - \polsub \geq \E_{M \sim \mathcal{D}} \left[ \V{s}{M}{\pi^*} \right] - 2\polsub - 3 \solvesub - \frac{\epsilon}{H}.
$$
This exactly demonstrates the claim in Eq.~\eqref{eqn:upper_bound_claim}, which as argued earlier, is sufficient to complete the proof of the theorem.

\subsection{Proof Of Lemma~\ref{lemma:upper_bound_helper_one}}
\label{app:upper_lem1}
\noindent Let $s \equiv s_t$ and recall that $s_1$ and $s_2$ denote the child states obtained by taking actions $a_1$ and $a_2$ respectively from state $s$. The assumption $\pi^*(s) = a_1$ is WLOG, since the case when $\pi^*(s) = a_2$ is entirely symmetric. Now consider any MDP $M$ and recall $\pi^*_M$ denotes an optimal policy for MDP $M$. There are two cases to consider. \\

\noindent For the first case, assume there exists $\pi^*_M$ such that $\pi^*_M(s) = a_1$. Then,
$$
\Rew_M(s, a_1) + \V{s_1}{M}{\pi^*_{M}} = \V{s}{M}{\pi^*_M} \geq \Rew_M(s, a_2) + \V{s_2}{M}{\pi^*_{M}} \geq \Rew_M(s, a_2) + \V{s_2}{M}{\pi^*_{M}} - \polsub,
$$
where we used the principle of Bellman optimality. \\

\noindent For the second case, assume there only exists $\pi^*_M$ such that $\pi^*_M(s) = a_2$. Then,
\begin{align*}
\Rew_M(s, a_1) + \V{s_1}{M}{\pi^*_M} &\geq \Rew_M(s, a_1) + \V{s_1}{M}{\pi^*} \\
&= \V{s}{M}{\pi^*} \\
&\geq \V{s}{M}{\pi^*_M} - \polsub \\
&= \Rew_M(s, a_2) + \V{s_2}{M}{\pi^*_{M}} - \polsub,
\end{align*}
where the first inequality follows from the optimality of $\pi^*_M$ for $M$, the equalities follow from the principle of Bellman optimality as well as the WLOG assumption that $\pi^*(s) = a_1$, and the second inequality follows from \strprox{} (b). \\

\noindent So in either case we are guaranteed that
$$
\Rew_M(s, a_1) + \V{s_1}{M}{\pi^*_M} \geq \Rew_M(s, a_2) + \V{s_2}{M}{\pi^*_{M}} - \polsub.
$$
Recall that for action $a$ leading to state $s'$ from state $s$, we have $Q_{i, a} = \Rew_{M_i}(s, a) + \approxV{s'}{M_i}$ by definition. We are then guaranteed by WIO that
$$
\Rew_{M_i}(s, a) + \V{s'}{M_i}{\pi^*_{M_i}} \geq Q_{i, a} \geq \Rew_{M_i}(s, a) + \V{s'}{M_i}{\pi^*_{M_i}} - \solvesub.
$$
Hence for any $M_i$ we must have
\begin{align*}
Q_{i, a_1} &= \Rew_{M_i}(s, a_1) + \approxV{s_1}{M_i} \\
&\geq \Rew_{M_i}(s, a_1) + \V{s_1}{M_i}{\pi^*_{M_i}} - \solvesub \\
&\geq \Rew_{M_i}(s, a_2) + \V{s_2}{M}{\pi^*_{M_i}} - \polsub - \solvesub \\
&\geq \Rew_{M_i}(s, a_2) + \approxV{s_2}{M_i} - \polsub - \solvesub \\
&= Q_{i, a_2} - \polsub - \solvesub.
\end{align*}
Averaging each side of the above inequality over $n$ completes the proof.

\subsection{Proof Of Lemma~\ref{lemma:upper_bound_helper_two}}
\label{app:upper_lem2}
Note that for action $a$ leading to state $s'$ from state $s$, we have $Q_{i, a} = \Rew_{M_i}(s, a) + \approxV{s'}{M_i}$ by definition. We are then guaranteed by WIO that
$$
\Rew_{M_i}(s, a) + \V{s'}{M_i}{\pi^*_{M_i}} \geq Q_{i, a} \geq \Rew_{M_i}(s, a) + \V{s'}{M_i}{\pi^*_{M_i}} - \solvesub.
$$
Hence we must have
$$
\frac{1}{n} \sum_{i=1}^n ( \Rew_{M_i}(s, a) + \V{s'}{M_i}{\pi^*_{M_i}} ) \geq \frac{1}{n} \sum_{i=1}^n Q_{i, a} \geq \frac{1}{n} \sum_{i=1}^n (\Rew_{M_i}(s, a) + \V{s'}{M_i}{\pi^*_{M_i}}) - \solvesub.
$$
By Hoeffding's bound and our choice of $n$, we are guaranteed that with probability at least $1 - \frac{\delta}{H}$ that
\begin{equation*}
\begin{split}
&\left \vert \frac{1}{n} \sum_{i=1}^n (\Rew_{M_i}(s, a_1) + \V{s_1}{M_i}{\pi^*_{M_i}}) - \E_{M \sim \mathcal{D}} \left[ \Rew_M(s, a_1) + \V{s_1}{M}{\pi^*_M} \right] \right \vert \leq \frac{\epsilon}{2H} \\
\text{ and } &\left \vert \frac{1}{n} \sum_{i=1}^n (\Rew_{M_i}(s, a_2) + \V{s_2}{M_i}{\pi^*_{M_i}} ) - \E_{M \sim \mathcal{D}} \left[ \Rew_M(s, a_2) + \V{s_2}{M}{\pi^*_M} \right] \right \vert \leq \frac{\epsilon}{2H}.
\end{split}
\end{equation*}
So for action $a$ leading to state $s'$ from state $s$, we can combine the previous two equations via the triangle inequality to obtain
\begin{align*}
&\left \vert \frac{1}{n} \sum_{i=1}^n Q_{i, a} - \E_{M \sim \mathcal{D}} \left[ \Rew_M(s, a) + \V{s'}{M}{\pi^*_M} \right] \right \vert \\
\leq &\left \vert \frac{1}{n} \sum_{i=1}^n (\Rew_{M_i}(s, a) + \V{s'}{M_i}{\pi^*_{M_i}}) - \frac{1}{n} \sum_{i=1}^n Q_{i, a} \right \vert \\
&+ \left \vert \frac{1}{n} \sum_{i=1}^n (\Rew_{M_i}(s, a) + \V{s'}{M_i}{\pi^*_{M_i}} ) - \E_{M \sim \mathcal{D}} \left[ \Rew_M(s, a) + \V{s'}{M}{\pi^*_M} \right] \right \vert \\
\leq &\solvesub + \frac{\epsilon}{2H}.
\end{align*}
This completes the proof.

\section{Near Tightness Of Theorem~\ref{thm:upper_bound}}
\label{app:prop_tight}
As discussed in Section~\ref{sec:upper_bound}, the $\polsub, \solvesub$ terms in the error bound of Theorem~\ref{thm:upper_bound} scale linearly with $H$. So when either $\polsub$ or $\solvesub$ is $\Omega(\frac{1}{H})$, then our bound becomes vacuous. It is natural to question whether this unfortunate scaling is due to a possible suboptimality of Algorithm~\ref{alg:main} or some looseness in our analysis. The following result provides a (partial) answer to this question.
\begin{proposition}
\label{prop:tight}
Let $n$ be the total query cost that any algorithm is allowed to use under WQM. For any $\solvesub \geq 0$, there exists $\mathcal{D}$ satisfying WIO with $\beta$ and \strprox{} with $\rewvar = \transvar = \polsub = 0$, such that the MDPs supporting $\mathcal{D}$ are deterministic and the following holds. Any (possibly randomized) algorithm will output (with probability at least $\frac{1}{2}$) a policy $\pi$ satisfying
$$
\E_{M \sim \mathcal{D}} \left[ \V{s_0}{M}{\pi} \right] \leq \max_{\pi'} \E_{M \sim \mathcal{D}} \left[ \V{s_0}{M}{\pi'} \right] - \frac{\solvesub H}{\log(50n)}.
$$
\end{proposition}
This result demonstrates that the dependency on $\solvesub$ given in the result of Theorem~\ref{thm:upper_bound} is tight to within a logarithmic factor in $H$, and cannot be improved beyond this logarithmic factor by a better algorithm or sharper analysis. It remains unclear whether the dependence on $\polsub$ is tight. Isolating this is more difficult, since in any construction $\polsub$ and $\solvesub$ become inherently intertwined, and it is unclear to us how to separate the dependencies on these two distinct parameters. We believe this is an interesting direction for future work. Intuitively the dependency on $\polsub$ seems tight --- roughly speaking it seems natural that if we are planning over $H$ timesteps, while using the structure induced by $\pi^*$ to guide our actions, then the suboptimality of $\polsub$ at each timestep leads to a total error blowup of $\polsub H$. Nevertheless, a formal proof (or otherwise) would be an interesting development. We now turn to proving Proposition~\ref{prop:tight}. \\

\noindent \begin{proof}
Fix any $\beta \geq 0$. It is sufficient to construct a single MDP $M$ with deterministic transitions, and construct an oracle $\oracle$ satisfying WIO with parameter $\beta$, and show that any algorithm will output a policy $\pi$ that satisfies
$$
\V{s_0}{M}{\pi} \leq \max_{\pi'} \V{s_0}{M}{\pi} - \frac{\beta H}{\log(50n)},
$$
with probability at least $\frac{1}{2}$. This is because we can define $\mathcal{D}$ to be the point mass on $M$, so that \strprox{} (a) and (b) are satisfied trivially with $\rewvar = \transvar = \polsub = 0$. We will define $M$ as a binary tree, where the states are nodes in the tree, and the (deterministic) actions are described by the edges connecting nodes, which immediately ensures that $M$ is deterministic. We will construct the following MDP $M$ by chunking the levels of the tree into blocks. Each block has length $\log(50n)$. To facilitate the definition of the reward function used to define $M$, we will first assign each state its optimal value, i.e. the value that one could get by following the optimal policy from that state. This will naturally allow us to later define rewards. We will assign these optimal values of the states in a sequential fashion, by considering each subtree rooted at some state on level $h_k = k \log(50n)$, where $k$ is a nonnegative integer. \\

\noindent Start by considering level $h_0 = 0$. Then on level $h_1 = \log(50n)$, pick a single state uniformly at random, denote it $s_{h_1, s_0}$, and assign it value that satisfies $V^{s_{h_1, s_0}}(\pi^*) = V^{s_0}(\pi^*)$. We will call this state the special state for level $h_1$. Let all other states $s$ on level $h_1$ have value $V^s(\pi^*) = V^{s_0}(\pi^*) - \beta$. \\

\noindent Recursively define the values of the leaves below this level in an identical fashion. To be more concrete, do the following procedure for each state $s$ on level $h_1 = \log(50n)$. Consider all the states in the subtree of $s$ that lie on level $h_2 = 2\log(50n)$. Pick a single one of these states uniformly at random, denote it $s_{h_2, s}$, and assign it value that satisfies
$$
V^{s_{h_2, s}}(\pi^*) = V^{s}(\pi^*).
$$
We will call this state one of the special states for level $h_2$. And for all other states $s' \neq s_{h_2, s}$ in the subtree of $s$ that lie on level $h_2 = 2\log(50n)$, assign them each value that satisfies
$$
V^{s'}(\pi^*) = V^{s}(\pi^*) - \beta.
$$
In this fashion, we can assign values for every state that lies on a level $k \log(50n)$ where $k$ is a nonnegative integer. We assume without loss of generality that $\log(50n)$ evenly divides $H$. Of course, this immediately defines the values for all states in the tree. To actually define rewards which satisfies the structure induced by the values, fix $\V{s_0}{M}{\pi^*}$ to be any number, and then simply assign rewards to the leaves of the tree which agree with the assigned values of those leaves. The reward is zero for all states in the tree that are not leaves. \\

\noindent We now define the oracle $\oracle$. For any state $s$ in the levels $h$ satisfying $h_0 \leq h \leq h_1$, let $\oracle(s) = V^{s_0}(\pi^*) - \beta$. Similarly, consider any state $s$ on level $h_1$, and now consider any state $s' \neq s$ in the subtree rooted at $s$ such that $\text{level}(s') \leq h_2$. Then let $\oracle(s') = V^{s}(\pi^*) - \beta$. Recursively repeat this until we have defined $\oracle$ for each state in the tree. By definition, we have defined $\oracle$ so that it satisfies WIO with parameter $\beta$. \\

\noindent With this construction in hand, the proof is now straightforward to complete. It is sufficient to prove this in the case when an algorithm returns a deterministic policy (or path), since any stochastic policy is a randomization over deterministic paths. We claim that with probability at least $\frac{1}{2}$, the path outputted by the algorithm never intersects \textit{any} of the special states in the \textit{entire} tree. It is sufficient to prove this, because this implies that the path loses $\beta$ value for a total of $\frac{H}{\log(50n)}$ times, implying that its value is at most
$$
V^{s_0}_M(\pi^*) - \frac{\beta H}{\log(50n)}.
$$
By symmetry, it is clear that any algorithm which can identify even a single special state anywhere in the tree with non-trivial probability, can be used to identify the special state on level $h_1$. Hence, it is sufficient to prove that there is no algorithm which can identify the special state on level $h_1$ with non-trivial probability. Strengthen the query model (slightly) so that querying a state will return whether it is the special state on level $h_1$. But now, observe that we are only allowed to query $n$ states total. And to identify the special state on level $h_1$, the algorithm must query most of the states on level $h_1$, which is a total of
$$
2^{\log(50n)} = 50n
$$
states. This implies that it cannot identify the special state on level $h_1$ with probability at least $\frac{1}{2}$. As discussed above, this is sufficient to complete the proof.
\end{proof}

\section{Auxiliary Proof Details}
\label{app:aux_proof}
In this section, we prove Lemma~\ref{lem:simulation}, relying heavily on the treatment given in~\citep{kakade03}. Note that in our setting, we only use this result in the context of our lower bounds, and all our lower bounds have finite state-action spaces. So it is sufficient to prove the result assuming that the state-action space is finite.

Let $\tau$ denote a trajectory, and let $\Prob_1^{\pi}(\tau), \Prob_2^{\pi}(\tau)$ denote the probabilities of taking $\tau$ in MDPs $M_1, M_2$ respectively. Let $\Rew_1(\tau), \Rew_2(\tau)$ denote the rewards obtained by following $\tau$ in MDPs $M_1, M_2$ respectively. The penultimate step of the proof of Lemma 4.3 in~\citep{kakade03} shows that
\begin{equation}
\label{eqn:aux_proof_1}
\sum_{\tau} \left \vert \Prob_1^{\pi}(\tau) - \Prob_2^{\pi}(\tau) \right \vert \leq \transvar H.
\end{equation}
The definition of $\rewvar$ and a straightforward application of the triangle inequality reveals that
\begin{equation}
\label{eqn:aux_proof_2}
\left \vert \Rew_1(\tau) - \Rew_2(\tau) \right \vert \leq \sum_{t = 0, (s_t, a_t) \in \tau}^{H-1} \left \vert \Rew_{M_1}(s_t, a_t) - \Rew_{M_2}(s_t, a_t) \right \vert \leq \rewvar H.
\end{equation}
Again using the triangle inequality, we obtain that
\begin{align*}
\left \vert \V{s_0}{M_1}{\pi} - \V{s_0}{M_2}{\pi} \right \vert &= \left \vert \sum_{\tau} \left( \Prob_1^{\pi}(\tau) \Rew_1(\tau) - \Prob_2^{\pi}(\tau) \Rew_2(\tau) \right) \right \vert \\
&\leq \left \vert \sum_{\tau} \left( \Prob_1^{\pi}(\tau) \Rew_1(\tau) - \Prob_1^{\pi}(\tau) \Rew_2(\tau) \right) \right \vert + \left \vert \sum_{\tau} \left( \Prob_1^{\pi}(\tau) \Rew_2(\tau) - \Prob_2^{\pi}(\tau) \Rew_2(\tau) \right) \right \vert \\
&\leq \sum_{\tau} \Prob_1^{\pi}(\tau) \left \vert \Rew_1(\tau) - \Rew_2(\tau) \right \vert + \left \vert \sum_{\tau} \left( \Prob_1^{\pi}(\tau) \Rew_2(\tau) - \Prob_2^{\pi}(\tau) \Rew_2(\tau) \right) \right \vert \\
&\leq \rewvar H + \left \vert \sum_{\tau} \left( \Prob_1^{\pi}(\tau) \Rew_2(\tau) - \Prob_2^{\pi}(\tau) \Rew_2(\tau) \right) \right \vert,
\end{align*}
where we used the result of Eq.~\eqref{eqn:aux_proof_2}. Furthermore,
\begin{align*}
\left \vert \V{s_0}{M_1}{\pi} - \V{s_0}{M_2}{\pi} \right \vert &\leq \rewvar H + \left \vert \sum_{\tau} \left( \Prob_1^{\pi}(\tau) \Rew_2(\tau) - \Prob_2^{\pi}(\tau) \Rew_2(\tau) \right) \right \vert \\
&\leq \rewvar H + \sum_{\tau} \left \vert \Prob_1^{\pi}(\tau)  - \Prob_2^{\pi}(\tau) \right \vert \left \vert \Rew_2(\tau) \right \vert \\
&\leq \rewvar H + \sum_{\tau} \left \vert \Prob_1^{\pi}(\tau)  - \Prob_2^{\pi}(\tau) \right \vert \\
&\leq \rewvar H + \transvar H,
\end{align*}
where we used the result of Eq.~\eqref{eqn:aux_proof_1} and also our assumption that the reward of any trajectory is always upper bounded by one. This completes the proof.

\bibliographystyle{alpha}
\bibliography{references_generalizable_rl}

\end{document}

%% file: commands.tex
\usepackage{amsmath,amssymb,graphicx,xcolor,mathtools,nicefrac}
\usepackage[shortlabels]{enumitem}
\newtheorem{theorem}{Theorem}
\newtheorem{lemma}{Lemma}

\newtheorem{corollary}{Corollary}
\newtheorem{proposition}{Proposition}

\newtheorem{condition}{Condition}
\newtheorem{property}{Property}

\usepackage{thmtools,thm-restate}

\newenvironment{proof}{{\bf Proof.}}{$\Box$}
\usepackage{hyperref}

\usepackage{amsmath,amssymb}
\usepackage{verbatim,float,url}
\usepackage{graphicx,subfigure,psfrag}
\usepackage[numbers]{natbib}
\usepackage{xcolor}
\usepackage{microtype}
\usepackage{xparse}
\usepackage{xspace}
\usepackage{tikz,textcomp,thmtools,nameref,cleveref}
\usetikzlibrary{positioning}
\usepackage{tikz-qtree}

\usepackage{algorithm, algorithmic}

\usepackage[font=small]{caption}

\newcommand*\circled[1]{\tikz[baseline=(char.base)]{
            \node[shape=circle,draw,inner sep=2pt] (char) {#1};}}

\newcommand{\strprox}{Strong Proximity}
\newcommand{\weakprox}{Weak Proximity}

\DeclareMathOperator*{\argmax}{argmax}

\DeclareMathOperator{\E}{\mathbb{E}}

\DeclareMathOperator{\R}{\mathbb{R}}
\DeclareMathOperator{\Prob}{\mathbb{P}}

\DeclareMathOperator{\StateSet}{\mathcal{S}}
\DeclareMathOperator{\ActSet}{\mathcal{A}}
\DeclareMathOperator{\Trans}{\mathcal{T}}

\DeclareMathOperator{\TvDist}{\mathbb{TV}}

\DeclareMathOperator{\rewvar}{\xi_r}
\DeclareMathOperator{\transvar}{\xi_{tr}}

\newcommand{\qone}{q_{\mathcal{D}}}

\newcommand{\Rew}{R}
\newcommand{\oracle}{\widehat{V}}

\DeclareMathOperator{\Mtest}{M_{test}}

\newcommand{\V}[3]{%
V^{{#1%
}}_{{#2%
}}({#3})
}%

\newcommand{\approxV}[2]{%
\widehat{V}^{{#1%
}}_{{#2%
}}}

\DeclareMathOperator{\polsub}{\alpha}
\DeclareMathOperator{\solvesub}{\beta}

\DeclarePairedDelimiterX{\infdivx}[2]{(}{)}{
  #1\;\delimsize|\delimsize|\;#2
}

%% file: rev_3_generalizable_rl.bbl
\newcommand{\etalchar}[1]{$^{#1}$}
\begin{thebibliography}{WCYW20}

\bibitem[AKY20]{agarwal20genmodel}
Alekh Agarwal, Sham Kakade, and Lin~F. Yang.
\newblock Model-based reinforcement learning with a generative model is minimax
  optimal.
\newblock In {\em Proceedings of the Conference on Learning Theory}, 2020.

\bibitem[AMCB21]{agarwal21contrastive}
Rishabh Agarwal, Marlos~C. Machado, Pablo~Samuel Castro, and Marc~G Bellemare.
\newblock Contrastive behavioral similarity embeddings for generalization in
  reinforcement learning.
\newblock In {\em International Conference on Learning Representations}, 2021.

\bibitem[AMK12]{azar12}
Mohammad~Gheshlaghi Azar, R\'{e}mi Munos, and Hilbert~J. Kappen.
\newblock On the sample complexity of reinforcement learning with a generative
  model.
\newblock In {\em Proceedings of the International Conference on Machine
  Learning}, 2012.

\bibitem[AN05]{abbeel05}
Pieter Abbeel and Andrew~Y. Ng.
\newblock Exploration and apprenticeship learning in reinforcement learning.
\newblock In {\em Proceedings of the International Conference on Machine
  Learning}, 2005.

\bibitem[BL13]{brunskill13}
Emma Brunskill and Lihong Li.
\newblock Sample complexity of multi-task reinforcement learning.
\newblock In {\em Proceedings of the Conference on Uncertainty in Artificial
  Intelligence}, 2013.

\bibitem[BMPS20]{bertran20}
Martin Bertran, Natalia Martinez, Mariano Phielipp, and Guillermo Sapiro.
\newblock Instance-based generalization in reinforcement learning.
\newblock In {\em Advances in Neural Information Processing Systems}, 2020.

\bibitem[BT03]{brafman03}
Ronen~I. Brafman and Moshe Tennenholtz.
\newblock R-max - a general polynomial time algorithm for near-optimal
  reinforcement learning.
\newblock {\em Journal of Machine Learning Research}, 3:213--231, 2003.

\bibitem[Cas20]{castro20similar}
Pablo~Samuel Castro.
\newblock Scalable methods for computing state similarity in deterministic
  markov decision processes.
\newblock In {\em Proceedings of the AAAI Conference on Artificial
  Intelligence}, 2020.

\bibitem[CHHS20]{cobbe19procgen}
Karl Cobbe, Chris Hesse, Jacob Hilton, and John Schulman.
\newblock Leveraging procedural generation to benchmark reinforcement learning.
\newblock In {\em Proceedings of the International Conference on Machine
  Learning}, 2020.

\bibitem[CKH{\etalchar{+}}19]{cobbe19}
Karl Cobbe, Oleg Klimov, Chris Hesse, Taehoon Kim, and John Schulman.
\newblock Quantifying generalization in reinforcement learning.
\newblock In {\em Proceedings of the International Conference on Machine
  Learning}, 2019.

\bibitem[CNL{\etalchar{+}}19]{clavera18}
Ignasi Clavera, Anusha Nagabandi, Simin Liu, Ronald~S. Fearing, Pieter Abbeel,
  Sergey Levine, and Chelsea Finn.
\newblock Learning to adapt in dynamic, real-world environments through
  meta-reinforcement learning.
\newblock In {\em International Conference on Learning Representations}, 2019.

\bibitem[CP10]{castro10transfer}
Pablo~Samuel Castro and Doina Precup.
\newblock Using bisimulation for policy transfer in mdps.
\newblock In {\em Proceedings of the AAAI Conference on Artificial
  Intelligence}, 2010.

\bibitem[DKWY20]{du20}
Simon~S. Du, Sham~M. Kakade, Ruosong Wang, and Lin~F. Yang.
\newblock Is a good representation sufficient for sample efficient
  reinforcement learning?
\newblock In {\em International Conference on Learning Representations}, 2020.

\bibitem[DLMW20]{du20detfunc}
Simon~S Du, Jason~D Lee, Gaurav Mahajan, and Ruosong Wang.
\newblock Agnostic q-learning with function approximation in deterministic
  systems: Near-optimal bounds on approximation error and sample complexity.
\newblock In {\em Advances in Neural Information Processing Systems}, 2020.

\bibitem[DLWZ19]{du19shiftoracle}
Simon~S Du, Yuping Luo, Ruosong Wang, and Hanrui Zhang.
\newblock Provably efficient q-learning with function approximation via
  distribution shift error checking oracle.
\newblock In {\em Advances in Neural Information Processing Systems}, 2019.

\bibitem[FAL17]{finn17}
Chelsea Finn, Pieter Abbeel, and Sergey Levine.
\newblock Model-agnostic meta-learning for fast adaptation of deep networks.
\newblock In {\em Proceedings of the International Conference on Machine
  Learning}, 2017.

\bibitem[FMB18]{farebrother18}
Jesse Farebrother, Marlos~C. Machado, and Michael Bowling.
\newblock Generalization and regularization in {DQN}.
\newblock {\em arXiv preprint arxiv:1810.00123}, 2018.

\bibitem[FMO20]{fallah20}
Alireza Fallah, Aryan Mokhtari, and Asuman Ozdaglar.
\newblock Provably convergent policy gradient methods for model-agnostic
  meta-reinforcement learning.
\newblock {\em arXiv preprint arxiv:2002.05135}, 2020.

\bibitem[FPP04]{ferns04}
Norm Ferns, Prakash Panangaden, and Doina Precup.
\newblock Metrics for finite markov decision processes.
\newblock In {\em Proceedings of the Conference on Uncertainty in Artificial
  Intelligence}, 2004.

\bibitem[FYY19]{feng19}
Fei Feng, Wotao Yin, and Lin~F. Yang.
\newblock Does knowledge transfer always help to learn a better policy?
\newblock {\em arXiv preprint arxiv:1912.02986}, 2019.

\bibitem[GHLL17]{gu17}
Shixiang Gu, Ethan Holly, Timothy Lillicrap, and Sergey Levine.
\newblock Deep reinforcement learning for robotic manipulation with
  asynchronous off-policy updates.
\newblock In {\em Proceedings of the IEEE International Conference on Robotics
  and Automation}, 2017.

\bibitem[HIB{\etalchar{+}}18]{henderson18}
Peter Henderson, Riashat Islam, Philip Bachman, Joelle Pineau, Doina Precup,
  and David Meger.
\newblock Deep reinforcement learning that matters.
\newblock In {\em Proceedings of the {AAAI} Conference on Artificial
  Intelligence}, 2018.

\bibitem[HPZ{\etalchar{+}}18]{haarnoja18}
T.~{Haarnoja}, V.~{Pong}, A.~{Zhou}, M.~{Dalal}, P.~{Abbeel}, and S.~{Levine}.
\newblock Composable deep reinforcement learning for robotic manipulation.
\newblock In {\em Proceedings of the IEEE International Conference on Robotics
  and Automation}, 2018.

\bibitem[Jia18]{jiang18}
Nan Jiang.
\newblock {PAC} reinforcement learning with an imperfect model.
\newblock In {\em Proceedings of the {AAAI} Conference on Artificial
  Intelligence}, 2018.

\bibitem[KK99]{kearns99}
Michael Kearns and Daphne Koller.
\newblock Efficient reinforcement learning in factored mdps.
\newblock In {\em Proceedings of the International Joint Conference on
  Artificial Intelligence}, 1999.

\bibitem[KKL03]{kakade03}
Sham Kakade, Michael Kearns, and John Langford.
\newblock Exploration in metric state spaces.
\newblock In {\em Proceedings of the International Conference on Machine
  Learning}, 2003.

\bibitem[KMN99]{kearns99generative}
Michael Kearns, Yishay Mansour, and Andrew~Y. Ng.
\newblock A sparse sampling algorithm for near-optimal planning in large markov
  decision processes.
\newblock In {\em Proceedings of the International Joint Conference on
  Artificial Intelligence}, 1999.

\bibitem[KS02]{kearns02}
Michael Kearns and Satinder Singh.
\newblock Near-optimal reinforcement learning in polynomial time.
\newblock {\em Machine Learning}, 49:209--232, 2002.

\bibitem[KS06]{kocsis06}
Levente Kocsis and Csaba Szepesv\'{a}ri.
\newblock Bandit based monte-carlo planning.
\newblock In {\em Proceedings of the European Conference on Machine Learning},
  2006.

\bibitem[LBC21]{lan21}
Charline~Le Lan, Marc~G. Bellemare, and Pablo~Samuel Castro.
\newblock Metrics and continuity in reinforcement learning.
\newblock {\em arXiv preprint arxiv:2102.01514}, 2021.

\bibitem[LG10]{lazaric10}
Alessandro Lazaric and Mohammad Ghavamzadeh.
\newblock Bayesian multi-task reinforcement learning.
\newblock In {\em Proceedings of the International Conference on Machine
  Learning}, 2010.

\bibitem[LSW20]{lattimore20featurerep}
Tor Lattimore, Csaba Szepesvari, and Gellert Weisz.
\newblock Learning with good feature representations in bandits and in {RL}
  with a generative model.
\newblock In {\em Proceedings of the International Conference on Machine
  Learning}, 2020.

\bibitem[M{\etalchar{+}}15]{mnih15}
Volodymyr Mnih et~al.
\newblock Human-level control through deep reinforcement learning.
\newblock {\em Nature}, 518:529--533, 2015.

\bibitem[NPH{\etalchar{+}}18]{nichol18}
Alex Nichol, Vicki Pfau, Christopher Hesse, Oleg Klimov, and John Schulman.
\newblock Gotta learn fast: {A} new benchmark for generalization in {RL}.
\newblock {\em arXiv preprint arxiv:1804.03720}, 2018.

\bibitem[PGK{\etalchar{+}}18]{packer18}
Charles Packer, Katelyn Gao, Jernej Kos, Philipp Kr{\"{a}}henb{\"{u}}hl,
  Vladlen Koltun, and Dawn Song.
\newblock Assessing generalization in deep reinforcement learning.
\newblock {\em arXiv preprint arxiv:1810.12282}, 2018.

\bibitem[RD19]{vanroy19}
Benjamin~Van Roy and Shi Dong.
\newblock Comments on the du-kakade-wang-yang lower bounds.
\newblock {\em arXiv preprint arxiv:1911.07910}, 2019.

\bibitem[RLTK17]{rajeswaran17}
Aravind Rajeswaran, Kendall Lowrey, Emanuel~V. Todorov, and Sham~M Kakade.
\newblock Towards generalization and simplicity in continuous control.
\newblock In {\em Advances in Neural Information Processing Systems}. 2017.

\bibitem[RZF{\etalchar{+}}19]{rakelly19}
Kate Rakelly, Aurick Zhou, Chelsea Finn, Sergey Levine, and Deirdre Quillen.
\newblock Efficient off-policy meta-reinforcement learning via probabilistic
  context variables.
\newblock In {\em Proceedings of the International Conference on Machine
  Learning}, 2019.

\bibitem[S{\etalchar{+}}17]{silver17}
David Silver et~al.
\newblock Mastering the game of go without human knowledge.
\newblock {\em Nature}, 550:354--359, 2017.

\bibitem[SJT{\etalchar{+}}20]{song20}
Xingyou Song, Yiding Jiang, Stephen Tu, Yilun Du, and Behnam Neyshabur.
\newblock Observational overfitting in reinforcement learning.
\newblock In {\em International Conference on Learning Representations}, 2020.

\bibitem[SPM20]{sonar20}
Anoopkumar Sonar, Vincent Pacelli, and Anirudha Majumdar.
\newblock Invariant policy optimization: Towards stronger generalization in
  reinforcement learning.
\newblock {\em arXiv preprint arxiv:2006.01096}, 2020.

\bibitem[SWW{\etalchar{+}}18]{sidford18}
Aaron Sidford, Mengdi Wang, Xian Wu, Lin Yang, and Yinyu Ye.
\newblock Near-optimal time and sample complexities for solving markov decision
  processes with a generative model.
\newblock In {\em Advances in Neural Information Processing Systems}, 2018.

\bibitem[WAJ{\etalchar{+}}21]{weisz2021queryefficient}
Gellert Weisz, Philip Amortila, Barnabas Janzer, Yasin Abbasi-Yadkori, Nan
  Jiang, and Csaba Szepesvari.
\newblock On query-efficient planning in mdps under linear realizability of the
  optimal state-value function.
\newblock {\em arXiv preprint arxiv:2102.02049}, 2021.

\bibitem[WCYW20]{wang20meta}
Lingxiao Wang, Qi~Cai, Zhuoran Yang, and Zhaoran Wang.
\newblock On the global optimality of model-agnostic meta-learning.
\newblock In {\em Proceedings of the International Conference on Machine
  Learning}, 2020.

\bibitem[WDYS20]{wang20}
Ruosong Wang, Simon~S. Du, Lin~F. Yang, and Ruslan Salakhutdinov.
\newblock On reward-free reinforcement learning with linear function
  approximation.
\newblock In {\em Advances in Neural Information Processing Systems}. 2020.

\bibitem[WVR13]{zheng13detfunc}
Zheng Wen and Benjamin Van~Roy.
\newblock Efficient exploration and value function generalization in
  deterministic systems.
\newblock In {\em Advances in Neural Information Processing Systems}, 2013.

\bibitem[WZXS19]{wang19}
Huan Wang, Stephan Zheng, Caiming Xiong, and Richard Socher.
\newblock On the generalization gap in reparameterizable reinforcement
  learning.
\newblock In {\em Proceedings of the International Conference on Machine
  Learning}, 2019.

\bibitem[YQH{\etalchar{+}}19]{yu19}
Tianhe Yu, Deirdre Quillen, Zhanpeng He, Ryan Julian, Karol Hausman, Chelsea
  Finn, and Sergey Levine.
\newblock Meta-world: A benchmark and evaluation for multi-task and meta
  reinforcement learning.
\newblock In {\em Proceedings of the Conference on Robot Learning}, 2019.

\bibitem[ZBP18]{zhang18}
Amy Zhang, Nicolas Ballas, and Joelle Pineau.
\newblock A dissection of overfitting and generalization in continuous
  reinforcement learning.
\newblock {\em arXiv preprint arxiv:1806.07937}, 2018.

\bibitem[ZMC{\etalchar{+}}21]{zhang21bisim}
Amy Zhang, Rowan~Thomas McAllister, Roberto Calandra, Yarin Gal, and Sergey
  Levine.
\newblock Learning invariant representations for reinforcement learning without
  reconstruction.
\newblock In {\em International Conference on Learning Representations}, 2021.

\end{thebibliography}
